\documentclass{uwstat518}

\usepackage[utf8]{inputenc} 
\usepackage[T1]{fontenc}    
\usepackage{hyperref}       
\usepackage{url}            
\usepackage{booktabs}       
\usepackage{amsfonts}       
\usepackage{nicefrac}       
\usepackage{microtype}      

\usepackage{amssymb}
\usepackage{subcaption}
\usepackage{multirow}
\usepackage{changepage}
\usepackage{epstopdf}
\usepackage[dvipsnames]{xcolor}
\usepackage{colortbl}
\definecolor{LightCyan}{rgb}{0.8,1,1}
\definecolor{darkorange}{rgb}{1.0, 0.55, 0.0}\usepackage{setspace}
\usepackage{booktabs}
\usepackage{array}

\newcolumntype{M}{>{\centering\arraybackslash}m{\dimexpr}}

\newcommand{\presec}{\vspace*{-0pt}}
\newcommand{\postsec}{\vspace*{-0pt}}
\newcommand{\presubsec}{\vspace*{-0pt}}
\newcommand{\postsubsec}{\vspace*{-0pt}}
\newcommand{\preeq}{\vspace*{-0pt}}
\newcommand{\posteq}{\vspace*{-0pt}}
\newcommand{\prepar}{\vspace*{-0pt}}
\newcommand{\preenum}{\vspace*{-0pt}}
\newcommand{\postenum}{\vspace*{-0pt}}

\newcommand{\argmin}{\operatornamewithlimits{argmin}}

\newcommand{\colvec}[1]{\begin{bmatrix}#1\end{bmatrix}}

\newcommand{\bb}[1]{\mathbf{#1}}

\newcommand{\veps}{\varepsilon}
\newcommand{\sumi}[2]{\sum_{{#1}=1}^{#2}}

\def\bal#1\eal{\begin{align*}#1\end{align*}}
\newcommand{\R}{\mathbb{R}}
\newcommand{\T}{\intercal}

\newcommand{\removed}[1]{}

\title{A Unified Framework for Long Range and Cold Start Forecasting of Seasonal Profiles in Time Series}

\author{Christopher Xie$^1$, Alex Tank$^2$, Alec Greaves-Tunnell$^2$, Emily Fox$^{1,2}$\\
$^1$Computer Science and Engineering,  $^2$Department of Statistics\\
University of Washington
}

\begin{document}

\maketitle

\begin{abstract}

Providing long-range forecasts is a fundamental challenge in time series modeling, which is only compounded by the challenge of having to form such forecasts when a time series has never previously been observed.  The latter challenge is the time series version of the \emph{cold-start} problem seen in recommender systems which, to our knowledge, has not been addressed in previous work. A similar problem occurs when a long range forecast is required after only observing a small number of time points --- a \emph{warm start} forecast. With these aims in mind, we focus on forecasting seasonal profiles---or \emph{baseline demand}---for periods on the order of a year in three cases: the long range case with multiple previously observed seasonal profiles, the cold start case with no previous observed seasonal profiles, and the warm start case with only a single partially observed profile. Classical time series approaches that perform iterated step-ahead forecasts based on previous observations struggle to provide accurate long range predictions; in settings with little to no observed data, such approaches are simply not applicable. Instead, we present a straightforward framework which combines ideas from high-dimensional regression and matrix factorization on a carefully constructed data matrix. Key to our formulation and resulting performance is leveraging (1) repeated patterns over fixed periods of time and across series, and (2) metadata associated with the individual series; without this additional data, the cold-start/warm-start problems are nearly impossible to solve. We demonstrate that our framework can accurately forecast an array of seasonal profiles on multiple large scale datasets.
\end{abstract}

\section{Introduction}
\presec

\begin{figure}[th]
  \centering
  \includegraphics[scale=0.5]{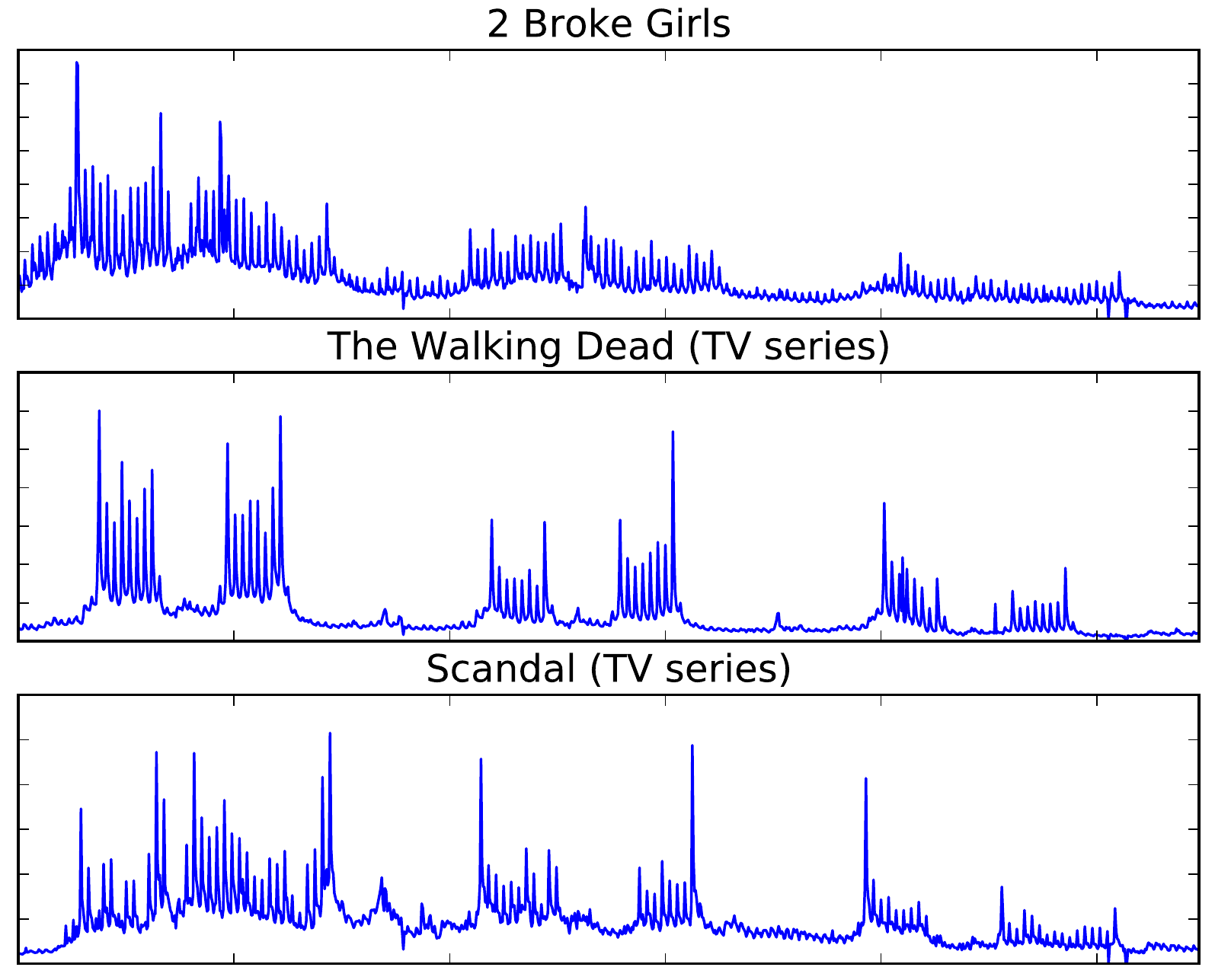}
  \caption{{\small Multiple seasonal profiles of Wikipedia page traffic of three entertainment TV shows over multiple years. These time series exhibit shared seasonality across the different series and across years, which relate to episode releases. We leverage such structure in order to provide accurate long-range forecasts.}}
  \label{fig:shared_seasonality}  
\end{figure}

Large collections of time series are now commonplace in many domains. Examples include environmental monitoring based on sensors at millions of locations, product demand and purchase curves for millions of products, and web traffic over time for billions of websites. A classical challenge in the time series domain is formulating long-range forecasts. For example, how does one forecast demand for a product in the coming year, so as to appropriately allocate inventory? 

The challenge of long-range forecasts is further compounded when having to make such a forecast of a previously \emph{unobserved} time series. For example, products on Amazon are introduced everyday and old products are taken down, new websites are created every minute and old ones disappear, and an autonomous vehicle detects newly seen pedestrians when they appear in its field-of-view. How can we provide a long-range forecast of demand/activity for this new product/website/pedestrian? We denote this problem as the \emph{cold-start forecasting} problem. To our knowledge, this important problem has not previously been directly addressed in the literature. Another related question of interest is: how can we refine our cold-start predictions after observing a few time points? We denote this problem variant as \emph{warm-start forecasting}. 

The long-range forecasting problem has traditionally been understudied, though see \cite{wen:2017,yu:2017} for recent approaches. The classical solution involves iteratively applying step-ahead predictions based on previously observed data. Such traditional time series methods include autoregressive (AR) based models (e.g. vector AR and seasonal AR) \cite{lutkepohl:2005}. Unfortunately, under slight model misspecification, iterating step-ahead forecasts to form long-range forecasts suffers from error accumulation when the length of the forecast window increases \cite{xia2011}. Furthermore, these prior methods fail to both capture relevant structure and scale to our high-dimensional settings of interest without making overly simplistic assumptions about the relationships between series, and are simply inapplicable on the cold/warm-start forecasting problems. In contrast to classical time series approaches, we propose a framework and demonstrate that it can accurately perform long-range forecasts and provide reliable cold/warm-start forecasts.

In order to produce a reliable forecast in data-scarce settings, we require that the time series exhibit some type of structure that makes them ``predictable''. 
Depending on the application, there are many features of time series one might be interested in predicting.  
For example, \cite{seeger2016bayesian} addressed the challenge of predicting intermittent demand of products. There is also the challenge of predicting a yearly trend component (e.g. growing popularity of a social media platform). Amongst the class of unpredictable structures are spikes in activity based on an unforeseeable event (e.g., a news event).  In this paper, our emphasis is on forecasting seasonal profiles (or patterns)---also known as \emph{baseline demand}---based on patterns observed both across fixed periods of time and series. While predicting the other types of structures are interesting problems as well, we limit our scope to seasonal profiles only. Methods predicting other structures can use our method as a component in a combined framework to forecast more than one of these structures.

Individual time series in a wide range of applications---such as product demand, website traffic, and sensor recordings---exhibit similar seasonal profiles and reactions to unobserved latent trends. For example, products related to skiing will see increases in demand during winter, and retail websites will peak close to Christmas. When considering many such individual series (i.e., a large-scale dataset), the result is a {\bf shared low-dimensional seasonal structure} (see Figure \ref{fig:shared_seasonality} for a real-world example of web page traffic related to TV shows). A number of authors have proposed applying matrix factorization techniques to collections of time series \cite{nguyen2014collaborative,sun2015time,Li:2015,Yu2015highdim,de2008analysis, agarwal:2017} to either provide retrospective analysis of shared latent structure or form short-term forecasts. However, these approaches ignore the cold-start challenge and do not directly handle long-range forecasts. 

In order to address the important cold-start challenge, we must leverage external features, which we denote as {\bf metadata}. Without external information describing the time series, forecasting without previously observed data is impossible. Luckily, it is now common that our large-scale time series have such associated metadata. For example, products have descriptions and user reviews, websites have content and network information, and sensors have locations and proximities to different points of interest. 
Prediction of time series curves from features, or \emph{covariates}, has traditionally been studied in the field of \emph{functional data analysis} (FDA) \cite{ramsay2005functional,morris2014functional}. 
Most uses of FDA relevant to our approach are only developed for a very small number of covariates---however, our metadata features are on the order of tens of thousands of dimensions.  More generally, such methods struggle to scale to the massive size of the datasets of interest to us. 

We harness the above time series characteristics---i.e., seasonality and relevant metadata---to jointly address the long-range, cold-start, and warm-start prediction challenges.  In particular, we propose a straightforward, computationally efficient framework that combines ideas of high-dimensional regression and matrix factorization. Our approach leverages a clever re-organization of the data matrix that amplifies the low-dimensional shared seasonality structure in the data. This allows us to tackle prediction of new, unobserved series using both high-dimensional metadata and shared seasonality structure, while being robust to missing data. The regression component of our model seeks to predict entire seasonal profiles from high dimensional metadata vectors, allowing us to form cold-start predictions in addition to long-range forecasts for previously seen items.  The matrix factorization component captures low-rank structure in the residuals unexplained by the regression and provides strengthened predictions in settings such as warm-start. Both ingredients leverage the shared seasonal profiles across years for a single item and across items by exploiting common relationships between (i) the metadata features and the seasonal profiles and (ii) the mean deviations captured in the residuals. 

Within our modular framework, one can propose different regression functions to tailor the framework to the dataset. To this end, we examine a few different regression functions motivated by real-world scenarios. For example, when the metadata is high-dimensional, we introduce a low-rank regression structure, and show this structure is crucial to our predictive performance. We explore the potential benefits and shortcomings for each of our considered cases in an analysis of two large data sets---page traffic for popular Wikipedia articles and Google flu trends for regions around the world---that shows our straightforward approach efficiently and effectively addresses the problem of forecasting seasonal profiles in a variety of challenging settings. To our knowledge, we report the first known results on the novel problems of cold-start and warm-start forecasting.

\postsec

\section{Background}
\presec

\label{sec:background}
\subsection{Matrix Factorization for Time Series}
\presubsec

\begin{figure*}
  \centering
  \includegraphics[width=5.0in]{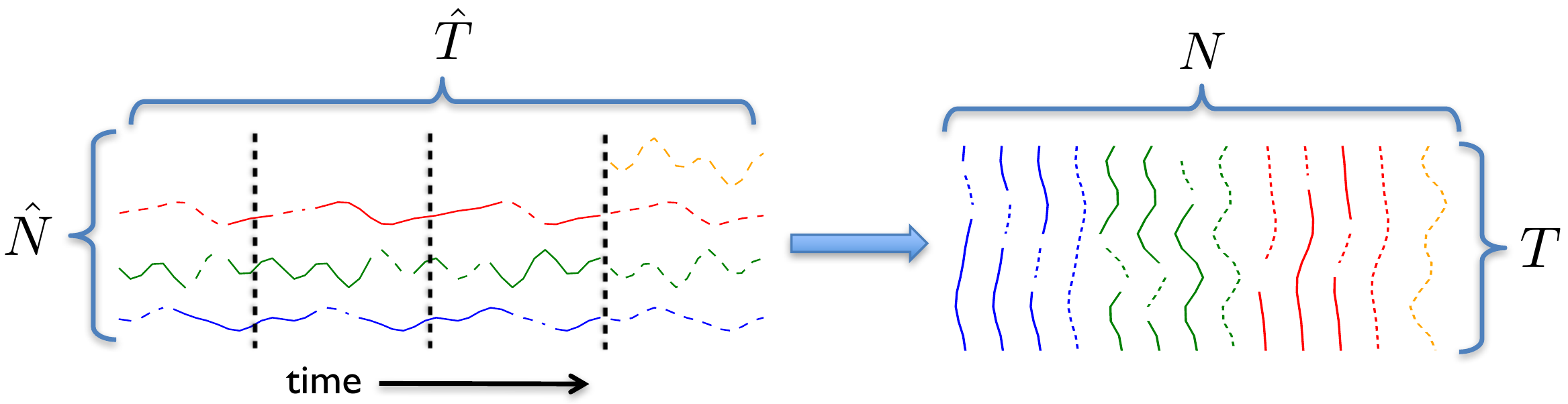}
  \caption{{\small Transformation from raw time series to our reorganized data matrix as described in Eq. (\ref{eq:datamat}) for three observed time series (blue, green, red) and one previously unobserved series (orange). The years are marked by the vertical thick dashed black lines. The solid colored lines indicate observed data points, and the dotted lines are missing values corresponding to the series having missing observations.}}
  \label{fig:stacking_matrix}  
\end{figure*}

Let $Y = [Y_1, \ldots, Y_N]$ denote a matrix of time series where each column, $Y_i$, represents a single length $T$ time series. Matrix factorization (MF) in this setting approximates $Y$ with a low-rank  product of two matrices $L \in \R^{T \times k}, R \in \R^{k \times N}$, where $k$ denotes the latent rank of the approximation. $L$ and $R$ may be computed by solving the general problem
\preeq
\begin{equation} 
\label{eq:mf}
\min_{L,R} \|Y - LR\|_F^2 + \lambda_L \mathcal{R}_L(L) + \lambda_R \mathcal{R}_R(R),
\end{equation}
\posteq
where $\mathcal{R}_L(L)$ and $\mathcal{R}_R(R)$ are regularization terms for $L$ and $R$, respectively. The columns of $L$ represent $k$ latent time series features and the rows of $R$ are the feature loadings for each series. If entries in $Y$ are missing, the learned latent series $L$ and loadings $R$ may be used to impute the missing values. Missing data imputation using matrix factorization is referred to as collaborative filtering \cite{koren2009matrix}.

Variations of Problem (\ref{eq:mf}) have appeared in the literature. 
For example, a state-of-the-art regularized temporal matrix factorization approach was developed in \cite{Yu2015highdim} where a penalty on $L$ encourages an autoregressive model of the latent time series factors. Smoothness of the latent factors across time may also be enforced via a Gaussian process prior \cite{nguyen2014collaborative}. Both \cite{sun2015time} and \cite{de2008analysis} perform non-negative matrix factorization on $Y$ where both $L ,R \geq 0$. This approach is used for \emph{interpretability} of the latent factors and factor loadings. Since our goal is prediction, we consider unrestricted $L$ and $R$ matrices.

\postsubsec
\subsection{Leveraging metadata in prediction}
\presubsec

Collaborative filtering is greatly improved by leveraging metadata \cite{koren2009matrix, agarwal2009regression}. For example, in the cold-start setting when a user has not rated any items, user metadata may be used to help predict ratings. Linear regression on the user and item metadata is added to the matrix factorization objective to obtain cold-start predictions \cite{pilaszy2009recommending,schein2002methods,gantner2010learning,agarwal2009regression}, and leads to improved performance. 

In the time series context, one may have metadata or features about each series. Prediction of entire time series curves from features has been classically studied using \emph{functional regression} \cite{ramsay2005functional}. Here one predicts the value of time series $i$ at time $j$ as a linear combination of the length-$m$ feature vectors $\phi_i$
\preeq
\begin{equation}
\label{tv_reg}
Y_{ji} = \sum_{k=1}^m W_{jk} \phi_{ki} + \veps_{ji},
\end{equation}
\posteq
where $W_{jk}$ is the $k^\text{th}$ coefficient at time $j$, $\phi_{ki}$ the $k^\text{th}$ covariate for series $i$, and $\veps_{ji}$ is mean zero noise. To share information across time, the coefficients $W_{jk}$ are assumed to vary smoothly. To enforce smoothness of the regression weights $W_j$ across time steps $j$ we utilize a basis expansion. In particular, we assume that each element of $W_j$ may be written as a linear combination of \emph{basis functions} evaluated at time step $j$. Specifically, we have that element $k$ of vector $W_j$, $W_{jk}$, is written as
 \begin{align}
     W_{jk} = \sum_{h=1}^K b_{h}(j) Q_{hk},
 \end{align}
 where $b_h(j)$ is the $h$th smooth basis function evaluated at time point $j$ and the $Q_{hk}$ are linear combination coefficients that are constant across time. The basis weights $Q_{hk}$ $\forall h,k$ fully describes the functional regression model. Under this parameterization, Eq. (\ref{tv_reg}) becomes
\begin{align}
Y_{ji} &= \sum_{k=1}^m \sum_{h=1}^K b_h(j) Q_{hk} \phi_{ki} + \veps_{ji}\\
&= B_j^\T Q \phi_i + \veps_{ji},
 \end{align}
where the basis function weights are collected into the matrix $Q$ and here $B_j$ is the vector with elements $B_j = \left(b_{1}(j) ,\ldots, b_{K}(j)\right)$.

In typical time-varying regression settings, the dimensionality of the predictors is small \cite{ramsay2005functional}. In contrast, we consider time series prediction and forecasting in settings where the available metadata is  large (e.g. on the order of tens of thousands). In this setting, the number of potential regression coefficients is also extremely large, $m \times T$. Below, we introduce methods to deal with such high dimensional metadata in time series prediction. Our approach marries ideas of time-varying regression as in Eq. (\ref{tv_reg}) with a matrix factorization decomposition of multiple time series as in Eq. (\ref{eq:mf}), but performed after a careful data matrix manipulation.

\postsec
\section{Forecasting of Seasonal Profiles} \label{sec:methods}
\presec

\subsection{Data Re-organization}
\presubsec

Consider a multivariate time series $\left[{Y}_{1}, \ldots, {Y}_{\hat{N}}\right]$ with number of series $\hat{N}$ and total time points $\hat{T}$, i.e. $Y_i \in R^{\hat{T}}$. In order to leverage repeated patterns across fixed periods and series, we re-organize the data matrix to treat each year of each time series as a column. 
Thus, we form a new data matrix
\preeq
\begin{equation} \hspace{-.05in}
{Y} = \colvec{Y_{1}^{1},\ldots,Y_{1}^{n_1}, \ldots,Y_{\hat{N}}^1,\ldots, Y_{\hat{N}}^{n_{\hat{N}}}}
\label{eq:datamat}
\end{equation}
\posteq
where $Y_{i}^{u}$ denotes the $u^{\text{th}}$ year of the $Y_i$, $n_i$ denotes the number of years of data for series $i$. $T$ denotes the number of observations per year (e.g. 52 for weekly data, 365 for daily data) and $N$ is the number of observed years, i.e. $N := \sum_{i=1}^{\hat{N}} n_i$. The reorganized matrix has dimensions $T \times N$, see Figure \ref{fig:stacking_matrix}. We perform all modeling with this data matrix. Note that we discriminate between the original time series $Y_i$ and the {\it seasonal profiles} (columns of $Y$ in Eq. \ref{eq:datamat})) $Y_i^u, u = 1, \ldots, n_i$ by the index $u$. While we assume that a year is the length of a seasonal period, any relevant length can be used.

Following our discussion on leveraging metadata, we assume that each individual time series $Y_{i}$ is accompanied by a metadata vector $\phi_i \in \R^m$. When performing the data re-organization in Eq. \ref{eq:datamat}, we simply copy $\phi_i$ such that each seasonal profile $Y_i^u, u = 1, \ldots, n_i$ is accompanied by the metadata vector $\phi_i$.

If each time series consists of multiple seasonal periods (i.e. $n_i > 1$), this re-organization amplifies the low-dimensional structure of $Y$. To see this, imagine that each time series $Y_i$ contains multiple seasonal profiles of a similar structure, e.g. 5 years of skiing-related products seeing increased product demand every winter. The seasonal structure that is repeated across seasonal profiles of a single time series becomes a low-dimensional structure shared across columns of the re-organized data matrix $Y$ presented in Eq. (\ref{eq:datamat}), effectively encouraging it to be low-rank. Thus, even if the original time series already  shared structures across time series, then the re-organization further adds more low-dimensional structure. Figure \ref{fig:stacking_matrix} demonstrates this with a cartoon example, while Figure \ref{fig:shared_seasonality} shows a real-world example that exhibits the desired structure.
\postsubsec

\subsection{Model Framework}
\presubsec

Given a matrix $Y$ as in Eq. (\ref{eq:datamat}), we develop a framework of models that leverage data-driven discovery of low-dimensional structure via matrix factorization with various approaches for incorporating high-dimensional metadata. More precisely, our framework consists of a multivariate regression component $f: \R^m \rightarrow \R^T$ which operates on a metadata vector $\phi \in \R^m$, and a matrix factorization component $LR$, where $L \in \mathbb{R}^{T \times k'}, R \in \mathbb{R}^{k' \times N}$. To simplify notation, we now refer to the $i^{\textrm{th}}$ column of $Y$ as $Y_i$. We model the generative process of $Y_i$ as
\begin{equation} \label{eq:framework}
Y_i = f(\phi_i) + LR_i + b + \veps_i
\end{equation}
where $\phi_i$ is the metadata vector for time series $i$, $R_i$ is the $i^{\textrm{th}}$ column of $R$, $b \in \R^T$ is an intercept term, and $\veps_i \sim N(0,\sigma^2 I)$. All parameters are learned from data (see Section \ref{subsec:model_learning}). We discuss the motivation of each component below. 

\subsubsection{Regression Component}

The regression function $f$ allows us to leverage high-dimensional metadata features in predicting seasonal profiles. This allows us to not only better inform predictions in the traditional long-range prediction challenge, but also to employ our framework in the novel cold-start setting (recall that without a component that leverages external metadata about each individual time series, cold-start forecasting is impossible). As our framework is quite general, there are many possible analytic forms that can be used for $f$. In practice, it is best to choose a function that is well-suited to the problem at hand. In our work, we explore a number of possible formulations for the regression term, as outlined below:

\removed{
\begin{table*}[t]
  \caption{Model Framework}
  \label{base-model-table}
  \centering
\begin{tabular}{|c|c|c|c|c|}
    \hline
     & Simple \removed{Regression} & Low-Rank \removed{Regression} & Functional \removed{Regression} & Neural Network \removed{Regression} \\ \hline
    No MF & $Y_i = \textcolor{blue}{W\phi_i} + \veps_i$ & $Y_i = \textcolor{blue}{HU\phi_i} + \veps_i$ & $Y_i = \textcolor{blue}{BQ\phi_i} + \veps_i$ & $Y_i = \textcolor{blue}{g_\psi(\phi_i)} + \veps_i$ \\ \hline
    MF & $Y_i = \textcolor{red}{LR_i} + \textcolor{blue}{W\phi_i} + \veps_i$ & $Y_i = \textcolor{red}{LR_i}+ \textcolor{blue}{HU\phi_i} + \veps_i$ & $Y_i = \textcolor{red}{LR_i} + \textcolor{blue}{BQ\phi_i} + \veps_i$ & $Y_i = \textcolor{red}{LR_i} + \textcolor{blue}{g_\psi(\phi_i)} + \veps_i$ \\ \hline
\end{tabular}
\end{table*}
}


\prepar
\paragraph{Approach 1: Low-rank regression}
The simplest approach is to introduce a multivariate linear regression. However, in our settings of interest, the metadata dimensionality is high and such an approach does not scale well (the number of parameters for such an approach is $mT$, with $m$ on the order of tens of thousands). Thus, we introduce a low-rank multivariate linear regression described by $f(\phi_i) = HU\phi_i$, with $H \in \mathbb{R}^{T \times k}, U \in \mathbb{R}^{k \times m}$. Here, $k$ is the low rank with $k \ll T, m$. The columns of $H$ can be interpreted as time-varying weights, while $U\phi_i$ represents a low-dimensional representation of the metadata features. This structure is not only beneficial due to computational reasons, but also because of the low-rank regression better models the low-dimensional structure $Y$ (which is amplified by Eq. (\ref{eq:datamat})) than a full regression. The columns of $H$ could potentially include latent time series representing increases in skiing-related product demand in winter and/or Christmas peaks for retail websites. In Section \ref{sec:experiments}, we show that this approach is crucial to our predictive performance.

\prepar
\paragraph{Approach 2: Smoothly varying latent factors}
When the data exhibits temporal regularity such as smoothness, we can enhance our long-range forecasts by restricting the nature of its evolution. We achieve this by utilizing functional regression: $f(\phi_i) = BQ\phi_i$ where $B \in \R^{T \times (K+3)}$ is a B-spline matrix consisting of $K$ knots, resulting in $K+3$ basis functions (no intercept), and $Q \in \R^{(K+3) \times m}$ is a matrix of weights for each of the $m$ functional coefficients. One can view these models as using the smoothness assumption to fix $H$ (in the low-rank regression described above) to a smooth basis matrix $B$ (note that we typically choose $K \ll T$). Note that this approach still exploits the assumption that $Y$ has a low-dimensional structure.

\prepar
\paragraph{Approach 3: Neural network}
While the above two approaches are naturally motivated by low-rank and smoothness assumptions, they both are linear functions. In order to allow for more flexibility in the learned regression function, we introduce an arbitrary function approximator in the form of a neural network which allows for non-linear feature-to-output mappings. However, these models typically require large amounts of data to train \cite{goodfellow2016deep}. 
\\

Importantly, the strength of prediction from the regression component depends on how predictive the metadata is at capturing seasonal features of the series. In particular, there must be shared correlations between features of the metadata across series and features of the seasonal profiles. As we see in our experiments (Section \ref{sec:experiments}), this is often the case in real-world datasets. However, when this is not the case, we further exploit the shared seasonality structure by adding a matrix factorization component (see Section \ref{subsubsec:mf_comp}).

Note that, as presented, our framework makes the assumption that the baseline mean profile, as a function of the metadata, is the same across years. This occurs since we assume $\phi_i$ is constant across years $u = 1, \ldots n_i$; for applications where metadata is available at each year, this may be relaxed to obtain changing baseline demand.

\subsubsection{Matrix Factorization Component}
\label{subsubsec:mf_comp}

The matrix factorization (MF) component further leverages the inherent low-dimensional shared seasonality structure in $Y$ by modeling the error residuals (i.e. $Y_i - f(\phi_i)$) as low rank. It essentially aims to capture extra structure in the residuals that is not captured by the regression component. For example, if the metadata cannot distinguish between skiing-related products that go on sale before winter and those that do not, then the regression component will not be able to reliably predict that such products will see demand increases before most other skiing-related products. In such a case, the MF component will be able to capture this, resulting in improved prediction performance. 

The MF component allows each time series to deviate from the mean seasonal profile provided by the regression component $f(\phi)$. Consider the TV show time series in Figure \ref{fig:shared_seasonality}; one can observe that each of the series has a common structure of weekly episode releases. However, the exact starts and ends of each batch of episodes differs slightly, which can be modeled by the MF component.

Similar to the low-rank regression approach, the columns of $L$ can be interpreted as latent time series. $R_i$, the $i^{\textrm{th}}$, column of $R$, can be interpreted as a latent factor (i.e., learned weights on the latent time series) for $Y_i$. Additionally, the MF structure makes our framework adept at handling missing data \cite{koren2009matrix}. We show empirical results to demonstrate this in Section \ref{sec:experiments}.

Note that it is also possible to consider models that only utilize the regression component without the matrix factorization, i.e. modeling the generating process as $Y_i = f(\phi_i) + b + \veps_i$. We denote these models as the ``non-MF'' models. Such a model can still provide forecasts for long-range, cold-start, and warm-start settings. However, as we show in our experiments (Section \ref{sec:experiments}), the matrix factorization component is useful in many of the prediction challenges, especially warm-start.

\postsubsec
\subsection{Model Learning}
\presubsec
\label{subsec:model_learning}

Our goal is to learn the parameters of these models while remaining robust to missing data. Let $\Omega \subseteq \{(j,i), j = 1, \ldots, T,\  i = 1, \ldots, N\}$ be the set of observed data indices of the data matrix $Y$ and $w$ be the parameters of the regression function $f$. To fit each of these models, we compute regularized maximum likelihood estimates of the parameters. Denoting $\|\cdot \|_F$ to be the Frobenius norm, this amounts to solving the optimization problem
\preeq
\begin{align}
\begin{split}
\argmin_{w,L,R,b}\ &\frac{1}{2N} \sum_{(j,i) \in \Omega} \left(Y_{ji} - [f( \phi_i )]_j  - L_j^\T R_i -b_j \right)^2\\
 + &\frac{\lambda_1}{2N}  \mathcal{R}_f(f) + \frac{\lambda_2}{2N}  \left(\|L\|_F^2 + \|R\|_F^2\right),
\end{split}
\end{align}
\posteq
where $\mathcal{R}_f$ is a regularization function, $L_j^\T$ is the $j^{\textrm{th}}$ row of $L$, and $b \in \R^T$ is a bias. As an example, for the low-rank regression + matrix factorization model, we use the squared Frobenius norm to regularize $H$ and $U$, resulting in the objective
\preeq
\begin{align}
\begin{split}
\argmin_{H,U,L,R,b}\ &\frac{1}{2N} \sum_{(j,i) \in \Omega}\left(Y_{ji} - H_j^\T U\phi_i  - L_j^\T R_i - b_j\right)^2\\
 + &\frac{\lambda_1}{2N}  \left( \|H\|_F^2 + \|U\|_F^2\right) + \frac{\lambda_2}{2N}  \left( \|L\|_F^2 +\|R\|_F^2\right).
\end{split}
\end{align}
\posteq
where $H_j^\T$ is the $j^{\textrm{th}}$ row of $H$. For the non-MF models we fix $L_j=0$ and $R_i=0$ for all $j,i$. We use mini-batch stochastic gradient descent (SGD) to learn the models. Calculating the gradient requires $O(N \max\{Tk', C\})$ complexity, where $C$ is the complexity of computing $f(\phi)$, e.g. for the low-rank regression model, $C=O(km + Tk)$.  Thus each iteration of SGD obtains $O(n\max\{Tk, C\})$ complexity where $n$ is the size of the mini-batch. Note that although $m$ can be on the order of tens of thousands, if $\phi$ is sparse (as in our experiments in Section \ref{sec:experiments}) with sparsity level $\tilde{m}$, i.e. $\tilde{m} = \left|\left\{j:[\phi]_j \neq 0\right\}\right|$, then we can replace $m$ with $\tilde{m}$ in the complexity calculation.

\postsec

\subsection{Forming Predictions}
\presubsec

We briefly discuss the how our framework generates forecasts below. In our framework, forecasting is equivalent to imputing the dotted lines in Figure \ref{fig:stacking_matrix} (more detail can be found in Section \ref{sec:experiments}).

\paragraph{Cold-start/Long-range} To forecast with our framework models (Eq. (\ref{eq:framework})) on the long-range and cold-start prediction challenges, we use the learned regression component only since the latent factors $R_i$ are uninformed by data and driven to zero by the optimization objective. Thus, given metadata $\phi_i$ of a new time series or previously observed time series (for long-range forecasting), the forecast is simply $f(\phi_i)$. This means that the forecast is fully dependent on the metadata. Recall that in the cold-start setting, it is nearly impossible to make a reliable forecast without using metadata (unless there are strong assumptions on the data). Prediction with non-MF models is also given by $f(\phi_i)$, which utilizes the same set of parameters for making predictions, although these parameters were learned differently (one in the presence of the MF term and the other not).

\paragraph{Warm-start/Missing data} For the warm-start challenge, the limited number of observations allows us to estimate the latent factor $R_i$ of the MF component, thus we are able to forecast using both the regression and MF components. The forecast is given by $f(\phi_i) + LR_i$. We show in Section \ref{sec:experiments} that this greatly improves over cold-start forecasting. We can also use both components for missing data imputation, which we show in Section \ref{sec:experiments}. For non-MF models, all predictions are made only with the regression component. 
\postsec
\section{Datasets} \label{sec:datasets}
\presec

\removed{
\begin{figure}[t!]
  \centering
  \includegraphics[width=\linewidth]{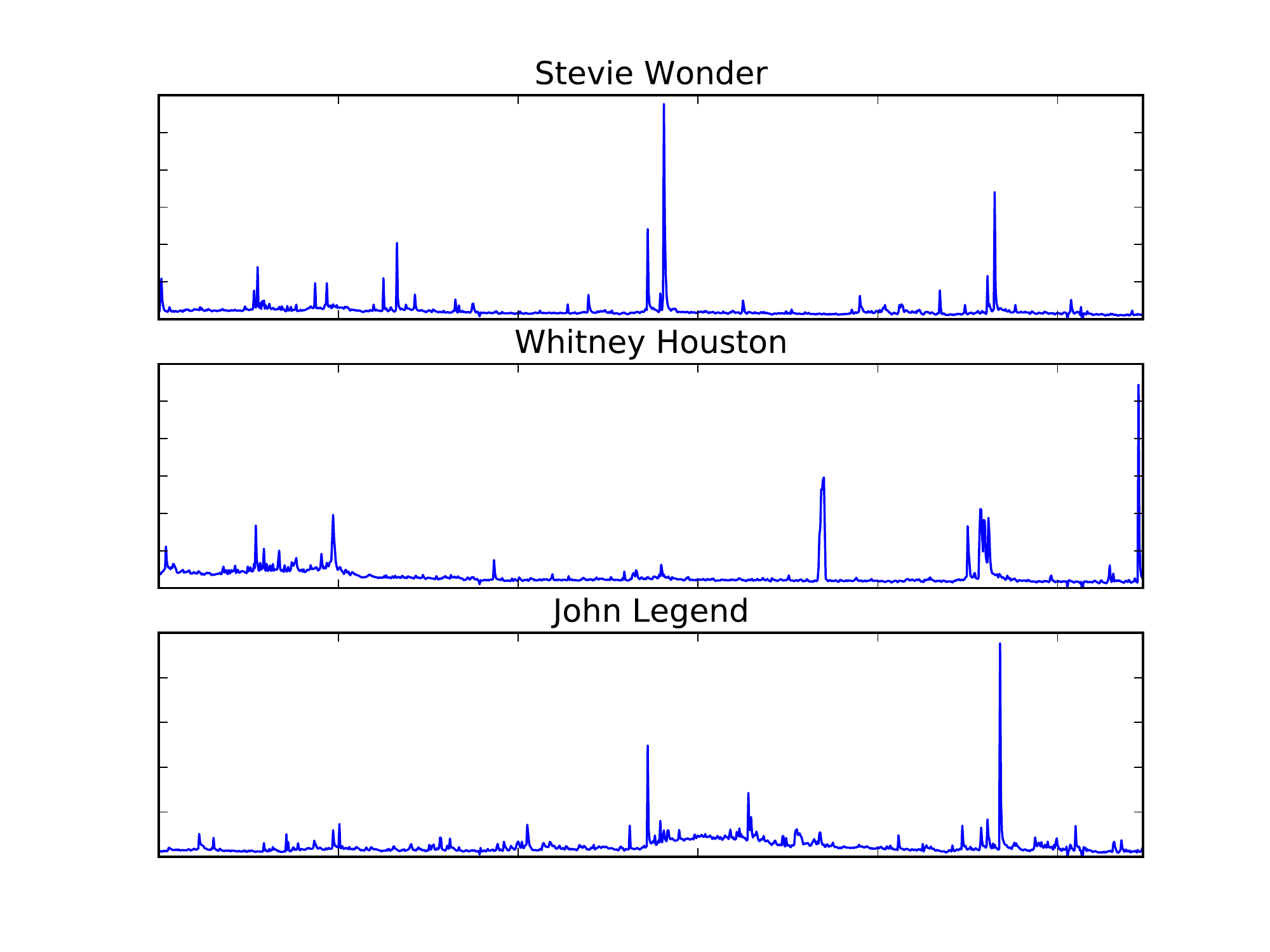}
  \caption{{\small Multiple seasonal profiles of Wikipedia page traffic of two famous musicians over multiple years from the \texttt{Wiki} dataset. Clearly, there is no shared low-dimensional structure, and there are irregular spikes, making forecasting difficult.}}
  \label{fig:spiky_series}  
\end{figure}
}

We explore our framework on all four prediction challenges arising in two very different datasets: Google Flu Trends (\texttt{Flu}) and Wikipedia page traffic (\texttt{Wiki}). The \texttt{Flu} dataset consists of weekly estimates of influenza rates in 311 worldwide regions~\cite{ginsberg2009detecting}.
To create our \texttt{Wiki} dataset, we scraped multiple years of daily page traffic data from 4031 of the most popular Wikipedia articles. A full description of the datasets can be found in Appendix \ref{appendix:experiment_details}.

For both datasets, we detrend the data using the method of \cite{cleveland1990stl} as we are focused on forecasting the seasonal profiles of the time series. The metadata is curated by scraping the relevant Wikipedia summaries---either from the geographic region's page (\texttt{Flu}) or of the popular page itself (\texttt{Wiki})---and calculating TF-IDF representations of the text after standard preprocessing, resulting in sparse metadata vectors. We standardize the time series within each dataset to put them on the same scale. Full details of the preprocessing can be found in Appendix \ref{appendix:experiment_details}.  Table~\ref{table:real_data_dims} summarizes the dimensionality of each dataset.  Note that the metadata dimensionality for \texttt{Flu} is much smaller than that of \texttt{Wiki}; this is due to the similarity and sparsity of vocabulary used in Wikipedia page summaries for geographic regions.  

The two datasets differ significantly in structure. The \texttt{Wiki} data is an example of a complex real-world dataset for which prediction is desirable but challenging in practice. The seasonal profiles are varied, often with strong periodicity at the weekly level, but it exhibits many erratic spikes corresponding to news events, movie releases, or other sources of transient disruption. Despite this, our framework is still able to discover complex seasonal patterns while avoiding corruption from the spikes, as seen in Section \ref{subsec:qualitative_analysis}. The \texttt{Flu} dataset exhibits a simpler and smoother seasonal structure with regions within countries and/or hemispheres sharing very similar yearly patterns. 

\begin{table}[t!]
  \caption{Data properties}
  \label{table:real_data_dims}
  \centering
\begin{tabular}{|c|c|c|}
    \hline
    & Google Flu & Wikipedia \\ \hline
    \# series ($\hat{N}$) & 311 & 4031 \\ \hline
    \# columns of $Y$ ($N$) & 3197 & 29093 \\ \hline
    period ($T$) & 52 & 365 \\ \hline
    metadata dim ($m$) & 3356 & 22193 \\ \hline
\end{tabular}
\end{table}

\postsec

\section{Experiments}
\label{sec:experiments}
\presec

We evaluate the performance of the proposed framework on all four prediction challenges for the two real world datasets discussed in Section \ref{sec:datasets}. The experimental setups for the prediction challenges are as follows:

\preenum
\begin{itemize}
\item[1)] \textbf{Long-Range Forecasting}: The training set includes multiple previous years and metadata. The test set contains the final year for each series. We remove 20\% of the training data uniformly at random to simulate missing data conditions present in many real-world situations. Our goal is to forecast this last year for all series (i.e., entire dotted blue, green, and red lines shown in the right hand side of Figure \ref{fig:stacking_matrix}).

\item[2)] \textbf{Cold-Start Forecasting}: The test set consists of entirely heldout time series and associated metadata; none of the series that comprise our training set share this metadata. As above, we also remove 20\% of training data points uniformly at random. Our goal is to forecast a year for new time series (i.e., entire dotted orange line in Figure \ref{fig:stacking_matrix}) solely from metadata.

\item[3)] \textbf{Warm-Start Forecasting}:  This setting is identical to the cold-start setting except that the first few time points of each time series in the test set are included in the training set. The MF models are able to use that small amount of information in order to learn an initial estimate the latent factors $R_i$ for stronger prediction. In our experimental setup, the training set includes the first two months (8 weeks or 60 days) of data for each time series in the test set and the goal is to forecast the remaining 10 months.

\item[4)] \textbf{Missing Data Imputation}: We also evaluate the ability of our framework to impute missing data. The test set consists of contiguous chunks of missing elements in the data matrix $Y$ representing, for example, sensors temporarily failing. The training set contains the remaining observations in $Y$ and all metadata. Our goal is to impute the missing elements (i.e., portions of dotted lines embedded in the solid lines in the right hand side of Figure \ref{fig:stacking_matrix}).

\end{itemize}
\postenum
For each experiment, we run 5-fold cross validation over a grid to select the model hyperparameters. Due to the non-convexity of some of our loss functions, we run random restarts; this computationally intensive process leads us to perform an approximation to full grid search over hyperparameters as described in Appendix \ref{appendix:experiment_details}.

\postsubsec

\postsubsec
\subsection{Quantitative Analysis}
\presubsec

\begin{table*}[t!]
  \centering
  \small
\begin{adjustwidth}{-0.7in}{}
\begin{tabular}{|c||c|c|c|c||c|c|c|c|}
	\hline
	 & \multicolumn{4}{c||}{Google Flu} & \multicolumn{4}{c|}{Wikipedia} \\ \hline
     & LR & CS & WS & MD & LR & CS & WS & MD  \\ \hline
    \textcolor{red}{avg-PY} & .583/.404 & -- & -- & -- & .380/.344 & -- & -- & -- \\    \hline
    \textcolor{red}{k-NN} & -- & .592/.359 & .512/.210 & -- & -- & .259/.185 & .264/.177 & -- \\ \hline
    \textcolor{red}{TRMF} & 1.15/.879 & -- & -- & {\bf .220}/{\bf .146} & .326/.268 & -- & -- & .255/.208 \\ \hline
    \textcolor{red}{AR} & 1.29/.958 & -- & -- & -- & 2e8/.278 & -- & -- & -- \\ \hline
    \textcolor{red}{MF alone} & -- & -- & -- & .432/.347 & -- & -- & -- & .219/.193 \\ \hline \hline
    \textcolor{darkorange}{Low-Rank Reg.} & {\bf .559}/{\bf .370} & .511/{\bf .311} & .467/.227 & .472/.381 & .211/.157 & {\bf .238}/{\bf .162} & .230/.156 & .233/.201 \\ \hline
    \textcolor{blue}{MF+Low-Rank Reg.} & {\bf .560}/.381 & .521/{\bf .311} & .452/.214 & .358/.286 & {\bf .207}/{\bf .153} & {\bf .237}/{\bf .161} & {\bf .223}/{\bf .153} & {\bf .210}/{\bf .183} \\ \hline
    \textcolor{blue}{MF+Functiona Reg.} & .582/.411 & .520/.376 & {\bf .424}/{\bf .194} & .372/.277 & .247/.190 & .259/.181 & .238/.168 & .239/.208 \\ \hline
    \textcolor{blue}{MF+Neural Network} & .561/.375 & {\bf .503}/.321 & .429/.243 & .415/.330 & .231/.182 & .251/.176 & .244/.173 & .243/.210 \\ \hline    
\end{tabular}
\end{adjustwidth}
\caption{APST\_MSE/APST\_MAE for each model. \textcolor{red}{Baselines} are in red, a \textcolor{darkorange}{regression-only (non-MF) model} is in orange, and \textcolor{blue}{matrix factorization models} are in blue. Each prediction challenge is abbreviated by its initials (e.g. LR indicates Long-Range). The best performance in each scenario is bolded, as well as close performers. The low-rank regression models perform quite well across all scenarios.}
\label{table:numbers}
\end{table*}

We evaluate our framework on both real world datasets using two error metrics. As baselines, we compare to two models that use step-ahead forecasts (only when applicable): Temporal Regularized Matrix Factorization (\texttt{TRMF}) \cite{Yu2015highdim}, which is a state-of-the-art method for time series forecasting, and univariate autoregressive (\texttt{AR}) models. We also consider the following simple baselines: for the long-range setting, we average all past years (\texttt{avg\_PY}); for the missing data setting, we compare to matrix factorization alone (\texttt{MF}); for the cold-start and warm-start settings, we use the $k$ nearest neighbors (\texttt{$k$-NN}) in (Euclidean) metadata space and perform a weighted average of the associated time series. The following is important to note when considering these baseline methods:
\begin{itemize}
	\item As we see in Table~\ref{table:numbers}, no single baseline is capable of handling all prediction scenarios of interest.
	\item Since the cold-start (or warm-start) forecasting problem has not previously been studied in the literature, there are no additional relevant baseline methods beyond our proposed $\texttt{k-NN}$ baseline.
	\item Since the \texttt{AR} baseline cannot handle missing data, we learn this model in the \emph{fully observed data setting} (rather than 20\% missing); that is, in Table~\ref{table:numbers}, \texttt{AR} is not handicapped by the challenging missing data structure considered by other methods. For \texttt{avg\_PY}, and \texttt{$k$-NN}, we only average over the observed values. \texttt{TRMF} and \texttt{MF} can naturally handle missing data.
\end{itemize}
Training details of our framework models and baselines can be found in Appendix \ref{appendix:experiment_details}.

\subsubsection{Evaluation Metrics}

We evaluate prediction performance on two metrics, and show superiority of our proposed framework on both. The metrics are: 1) {\it average per series thresholded MSE} (APST\_MSE) and 2) {\it average per series thresholded MAE} (APST\_MAE). To calculate APST\_MSE, we first calculate MSE per series. We then calculate the mean of the per series MSE to get a metric that weights each test series equally. Because the \texttt{Wiki} dataset contains many large magnitude spikes, we threshold the per series MSE values. This allows the metric to convey information about the seasonal profile while disregarding the irrelevant spikes. The equation for APST\_MSE is:
\begin{equation} \label{eq:per_series_thresholded_MSE}
\text{APST\_MSE}\left(Y_{\text{test}}, \hat{Y}\right) = \frac1N \sumi{i}{N} \frac{\sumi{j}{T} \bb{1}\{|Y_{ji}| \leq \rho \} \left (Y_{ji} - \hat{Y}_{ji} \right)^2}{\sumi{i}{T} \bb{1}\{|Y_{ji}| \leq \rho \}}
\end{equation}
where $Y_{\text{test}}$ is the test set time series and $\hat{Y}$ is the predicted values. APST\_MAE is calculated in the same way, but using MAE in place of MSE, which amounts to replacing $\left (Y_{ji} - \hat{Y}_{ji} \right)^2$ with $\left |Y_{ji} - \hat{Y}_{ji} \right|$ in Eq. (\ref{eq:per_series_thresholded_MSE}). For \texttt{Wiki}, we set $\rho = 2$, which restricts evaluation to observations less than two standard deviations from the mean. The \texttt{Flu} dataset is much more well-behaved, thus we evaluate using normal MSE. Note that when $\rho$ is set to be larger than the maximum value in the dataset, APST\_MSE (APST\_MAE) reduces to MSE (MAE).

Many time series prediction methods are evaluated using a percentage error metric such as mean absolute percentage error, which requires division by the magnitude of the ground truth time series. Such a metric is useful when predicting high magnitude time series (e.g. ad revenue for large companies). However, in our setting, our data is zero-meaned and normalized to put each time series on the same scale. Since many time series values are close to zero, dividing by the magnitude of the time series will result in unstable calculations.

\subsubsection{Results and Analysis}

Results from our proposed framework along with baselines can be found in Table \ref{table:numbers}. We found that in each setting except for the cold start setting on \texttt{Flu}, adding \texttt{MF} to our regression models either aids in prediction performance or admits comparable results. Thus, we clearly see the benefits of using \texttt{MF} to learn latent residual factors. We considered all regression models both with and without \texttt{MF}, and the results followed exactly as in the low-rank regression case shown in Table~\ref{table:numbers}. As such, to focus our discussion on the relevant comparisons, we simply provide the remaining regression options with \texttt{MF}.

\prepar
\paragraph{Long-range forecasts} 
All models in our framework outperform (or are comparable to) the baselines on both metrics. The step-ahead methods \texttt{AR} and \texttt{TRMF} perform poorly on this problem due to error propagation. In fact, on the \texttt{Wiki} dataset, the \texttt{AR} baseline estimated a nonstationary model that exploded APST\_MSE. On both datasets, the low-rank regression models perform best. Note that in the volatile \texttt{Wiki} dataset, the functional regression model suffers as the data is not smooth.

\prepar
\paragraph{Cold-start forecasts} 
In this setting, the step-ahead methods are not applicable. All of our framework models perform better than (or comparable to) the \texttt{$k$-NN} baseline on both datasets. On \texttt{Flu}, the neural network regression plus \texttt{MF} performs best on APST\_MSE while the low-rank models perform best on APST\_MAE. They also perform best on \texttt{Wiki} for both metrics. Our framework provides strong predictions despite never having seen any of the historical data. In Section \ref{subsec:qualitative_analysis}, we demonstrate that our framework truly does learn to forecast interesting seasonal patterns.

\prepar
\paragraph{Warm-start forecasts} All models in our framework outperform the \texttt{k-NN} baseline on APST\_MSE. For APST\_MAE, our models outperform \texttt{k-NN} on \texttt{Wiki}, but it performs decently well on the simple \texttt{Flu} dataset.
The gains from adding \texttt{MF} to low-rank regression are apparent, and this trend was observed over all regression choices (not shown). This speaks to the fact that \texttt{MF} is effectively learning latent residual factors from the few observed time points, leading to stronger prediction.  Interestingly, functional regression plus \texttt{MF} performs best on \texttt{Flu}.  This method leverages the smoothness of this data; for warm-start, the first few observations help inform that smooth seasonal profile. For the more volatile \texttt{Wiki}, low-rank plus \texttt{MF} performs best. Note that the APST\_MSE/APST\_MAE values are lower than cold-start forecasts for all models and datasets. This shows the efficacy of using the limited number of observations to learn an initial estimate of the latent factors of the MF component. 

\prepar
\paragraph{Missing data imputations}
In the smooth setting of \texttt{Flu}, \texttt{TRMF} performs very well; filling in large contiguous blocks of missing data is tough for \texttt{MF} alone, but enforcing an AR regularization \cite{Yu2015highdim} appears to help a lot since \texttt{Flu} is well-modeled by such an assumption. Indeed, \texttt{TRMF} was designed specifically for such scenarios. However, \texttt{Wiki} does not exhibit such a simple structure and \texttt{TRMF} performs worse compared to our class of models. In fact, all of our framework models outperform \texttt{TRMF} on \texttt{Wiki}, showcasing the efficacy of our proposed framework even when the data violates the traditional assumptions of most step-ahead forecasters. Low-rank plus \texttt{MF} again provides the best performance. In fact, it is a top performer amongst our framework models on \texttt{Flu} as well. 

\prepar
\paragraph{Summary} In addition to being able to straightforwardly handle each scenario, whereas no baseline method can, our proposed framework models outperform the task-specific baselines in almost every scenario on two different metrics.  Interestingly, the popular class of step-ahead time series methods struggle to provide competitive long-range forecasts, and are not immediately applicable to cold-start and warm-start scenarios.  \texttt{avg-PY} is a straightforward and simple alternative for forecasting the next year of a seasonal series (again not handling cold-start and warm-start), but also does not yield competitive performance, especially on the more challenging \texttt{Wiki} data.  The same holds for \texttt{k-NN} in the cold-start and warm-start contexts.  Finally, \texttt{MF} alone is a go-to technique for imputing missing values in large data matrices; however, we see gains from adding our regression components, especially in the more structured \texttt{Flu} data.  Overall, low-rank regression plus \texttt{MF} appears to be a robust and effective method across all considered scenarios, and dominates on the complex \texttt{Wiki} data by exploiting the shared seasonality structure of the data matrix $Y$, which is only magnified by our clever re-organization of the data matrix as outlined in Section \ref{sec:methods}. 


\postsubsec
\subsection{Qualitative Analysis}
\label{subsec:qualitative_analysis}
\presubsec

\begin{figure*}[t!]
\begin{center}

\begin{subfigure}[t]{0.31\linewidth}
\includegraphics[width=\linewidth, height=2.5cm]{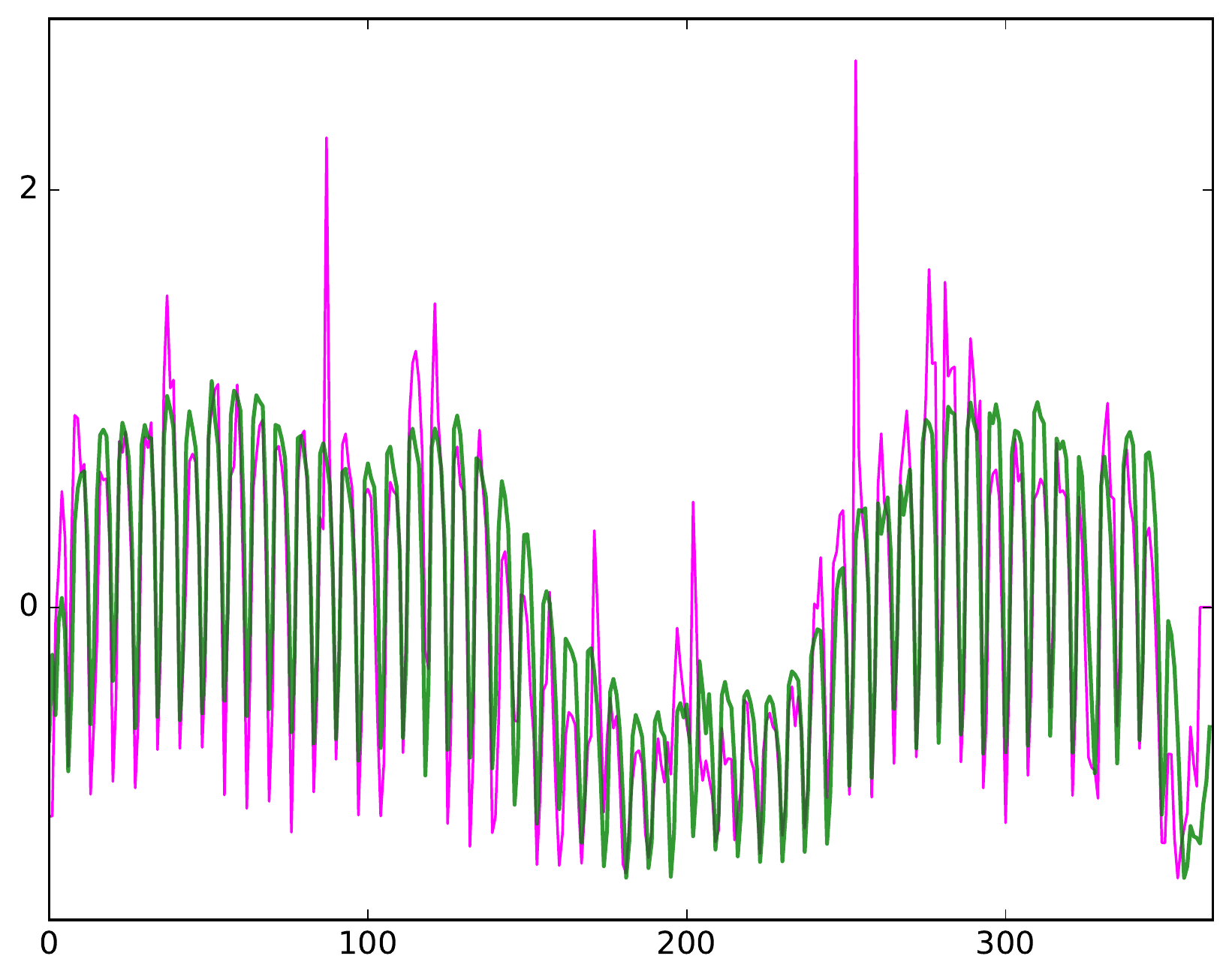}
\vspace{-.5cm}
\caption{Apollo 2014}
\label{fig:longrange_a}
\end{subfigure}
\hfill
\begin{subfigure}[t]{0.31\linewidth}
\includegraphics[width=\linewidth, height=2.5cm]{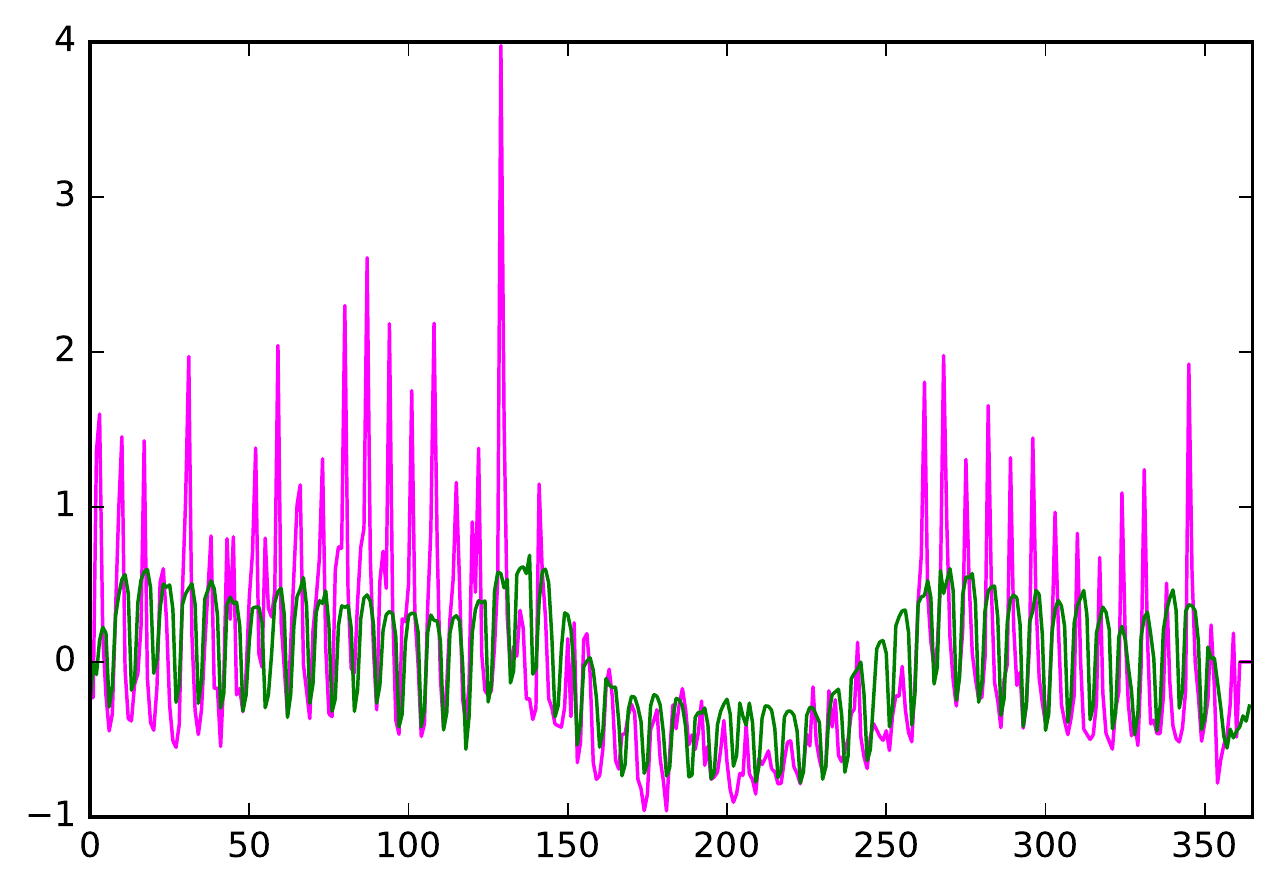}
\vspace{-.5cm}
\caption{NCIS: Los Angeles 2014}
\label{fig:longrange_b}
\end{subfigure}
\hfill
\begin{subfigure}[t]{0.31\textwidth}
\includegraphics[width=\linewidth, height=2.5cm]{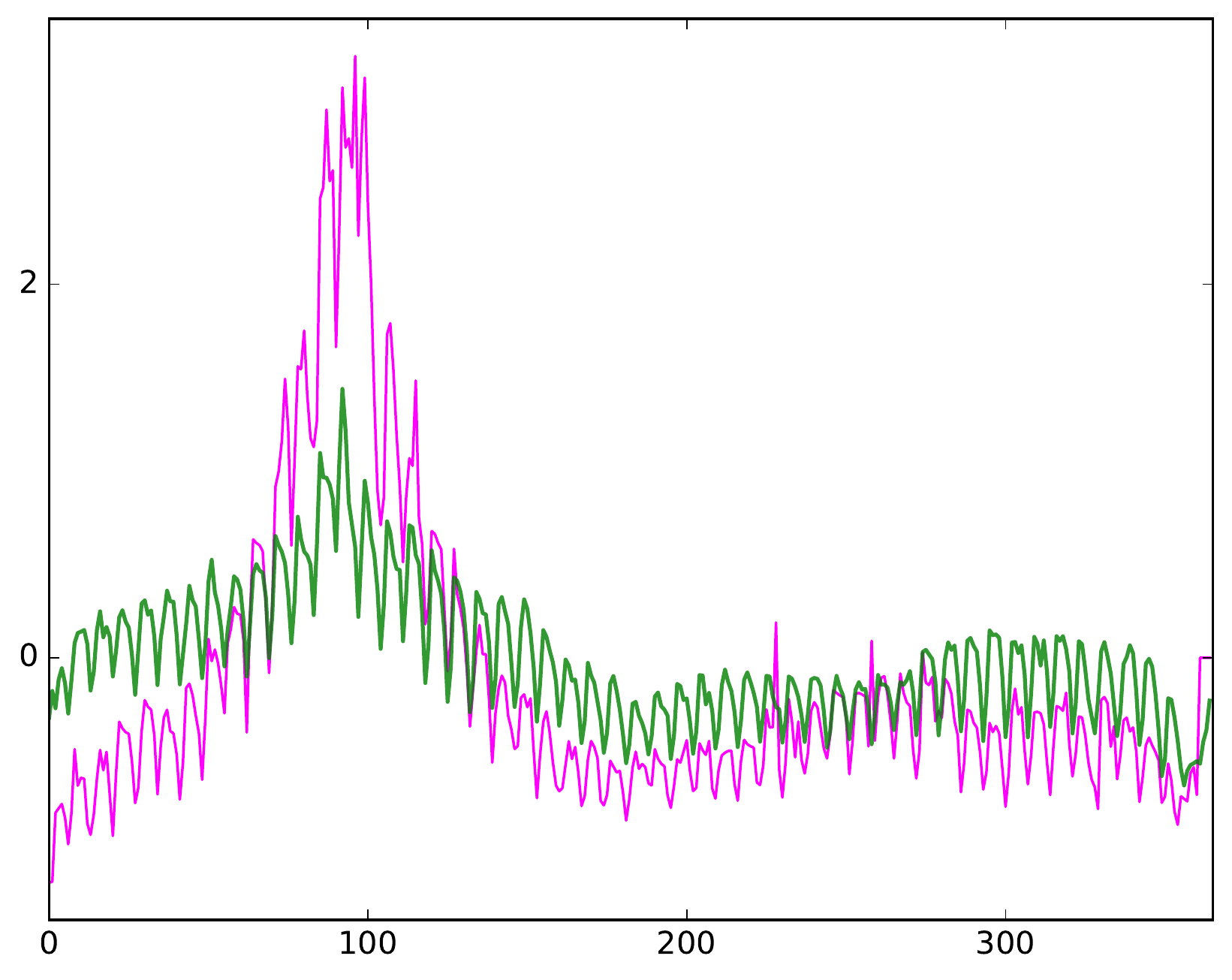}
\vspace{-.5cm}
\caption{Cherry Blossom 2014} 
\label{fig:longrange_c}
\end{subfigure}



\medskip
\vspace{-.2cm}

\begin{subfigure}[t]{0.31\linewidth}
\includegraphics[width=\linewidth, height=2.5cm]{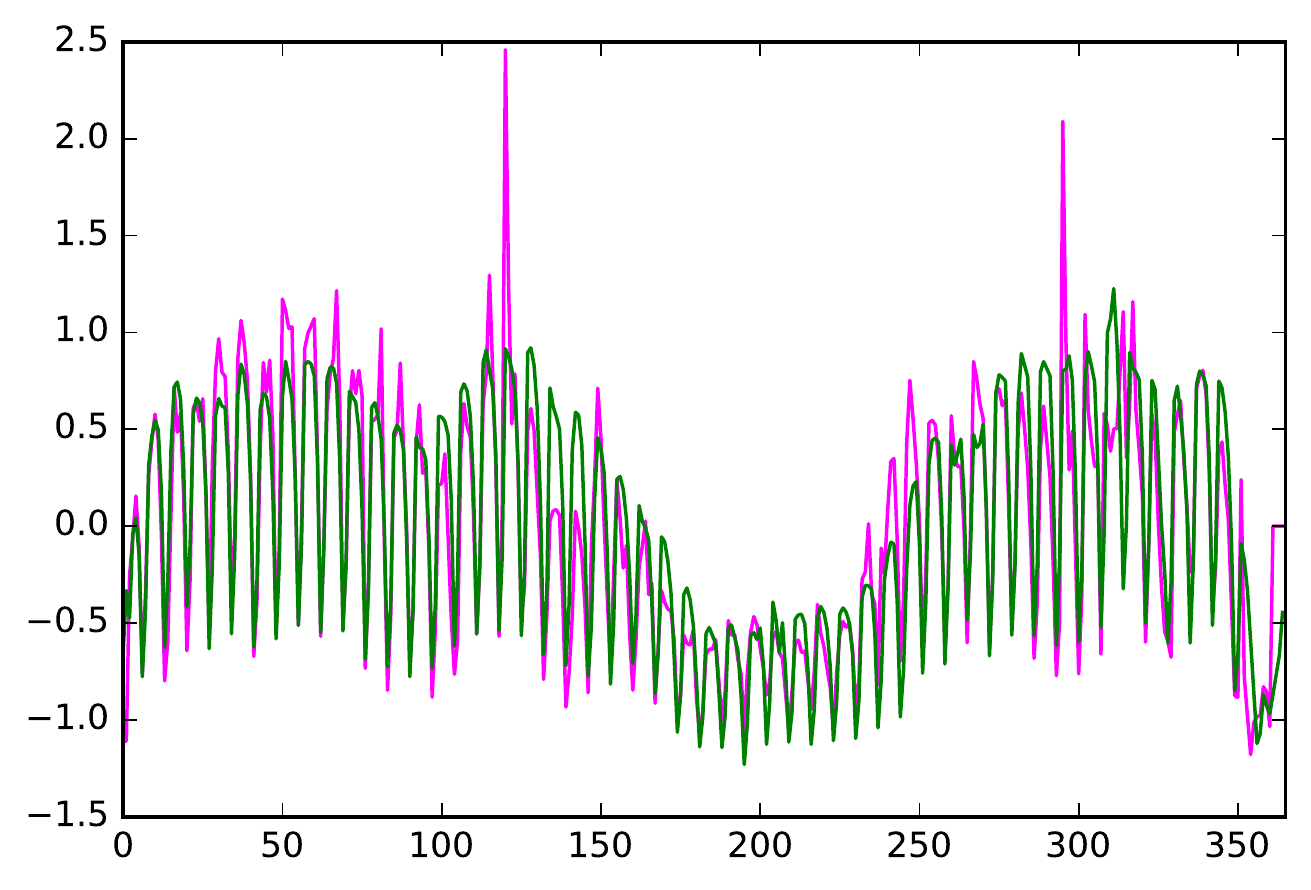}
\vspace{-.5cm}
\caption{Karl Marx 2014}
\label{fig:pure_a}
\end{subfigure}
\hfill
\begin{subfigure}[t]{0.31\textwidth}
\includegraphics[width=\linewidth, height=2.5cm]{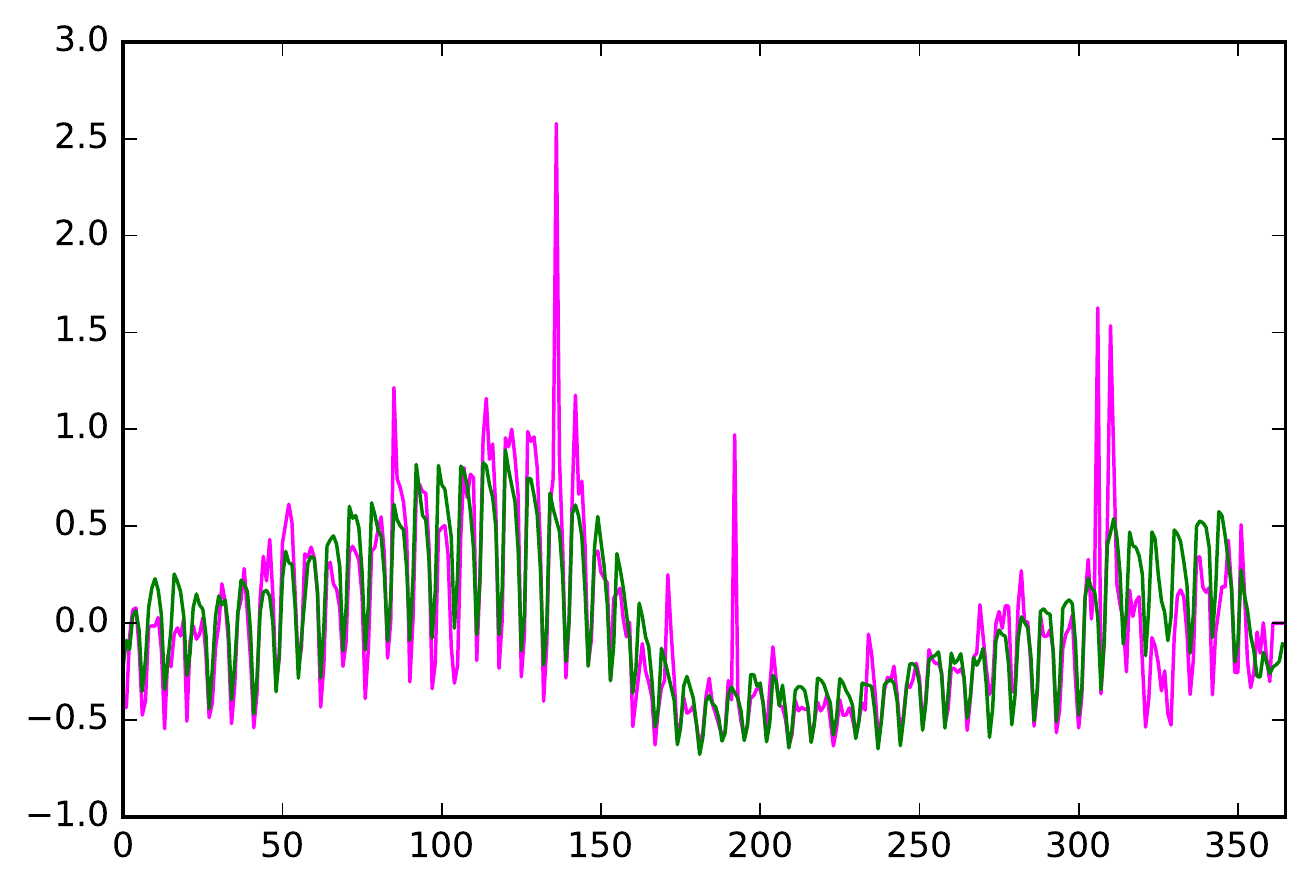}
\vspace{-.5cm}
\caption{Korean War 2014} 
\label{fig:pure_b}
\end{subfigure}
\hfill
\begin{subfigure}[t]{0.31\textwidth}
\includegraphics[width=\linewidth, height=2.5cm]{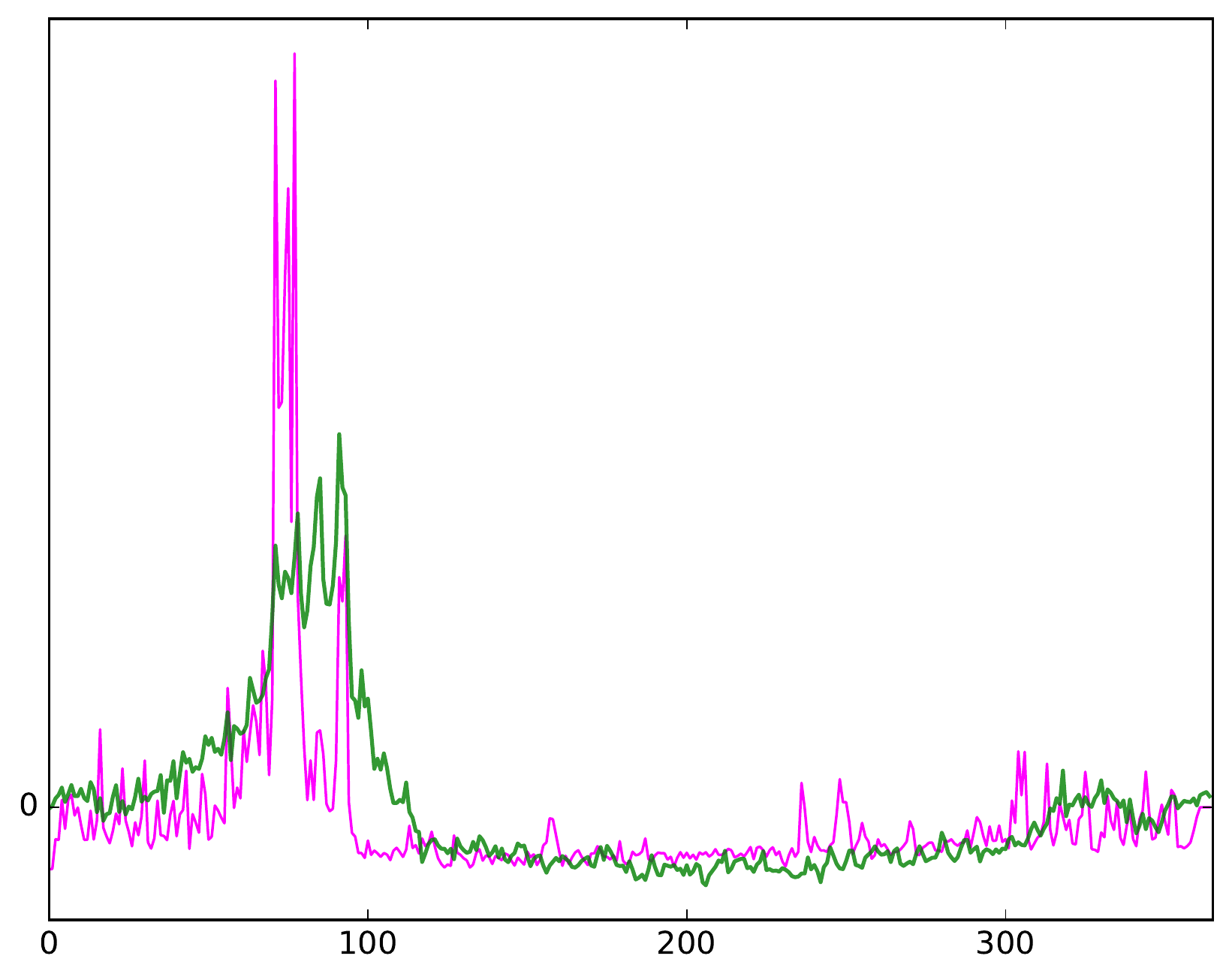}
\vspace{-.5cm}
\caption{Villanova Wildcats men's basketball 2014} 
\label{fig:pure_c}
\end{subfigure}
\vspace{-.2cm}
\caption{More predictions of the low-rank plus \texttt{MF} model on the \texttt{Wiki} dataset for long-range (top) and cold-start (bottom) challenges. The \textcolor{magenta}{observed time series} is in magenta and \textcolor{ForestGreen}{predictions} are overlayed in green. Best viewed in color on a computer screen.}
\label{fig:mflrr_predictions}
\end{center}
\end{figure*}

\begin{figure*}[t!]
\begin{center}

\begin{subfigure}[t]{0.31\linewidth}
\includegraphics[width=\linewidth, height=2.5cm]{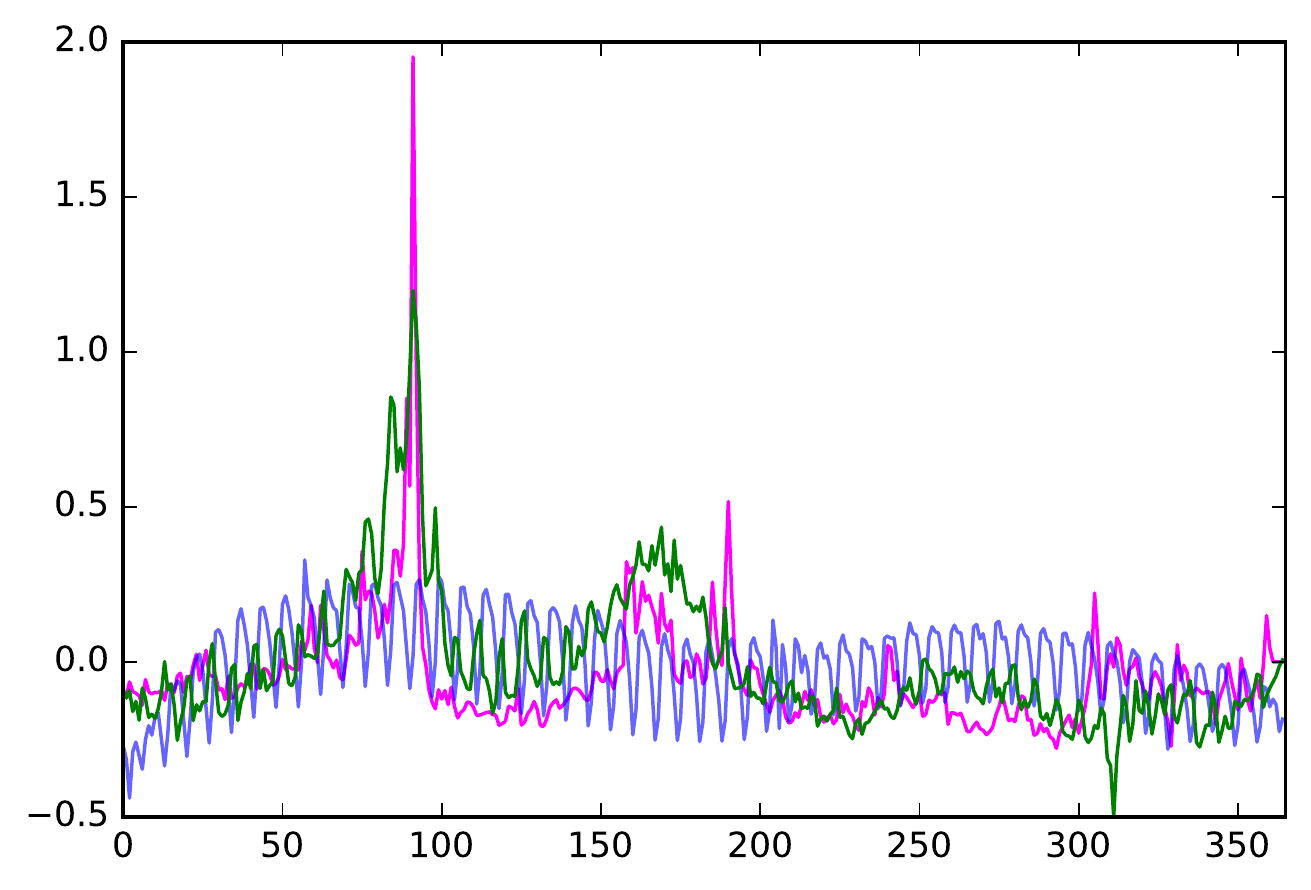}
\vspace{-.5cm}
\caption{Cricket World Cup 2014}
\label{fig:trmf_comp_a}
\end{subfigure}
\hfill
\begin{subfigure}[t]{0.31\linewidth}
\includegraphics[width=\linewidth, height=2.5cm]{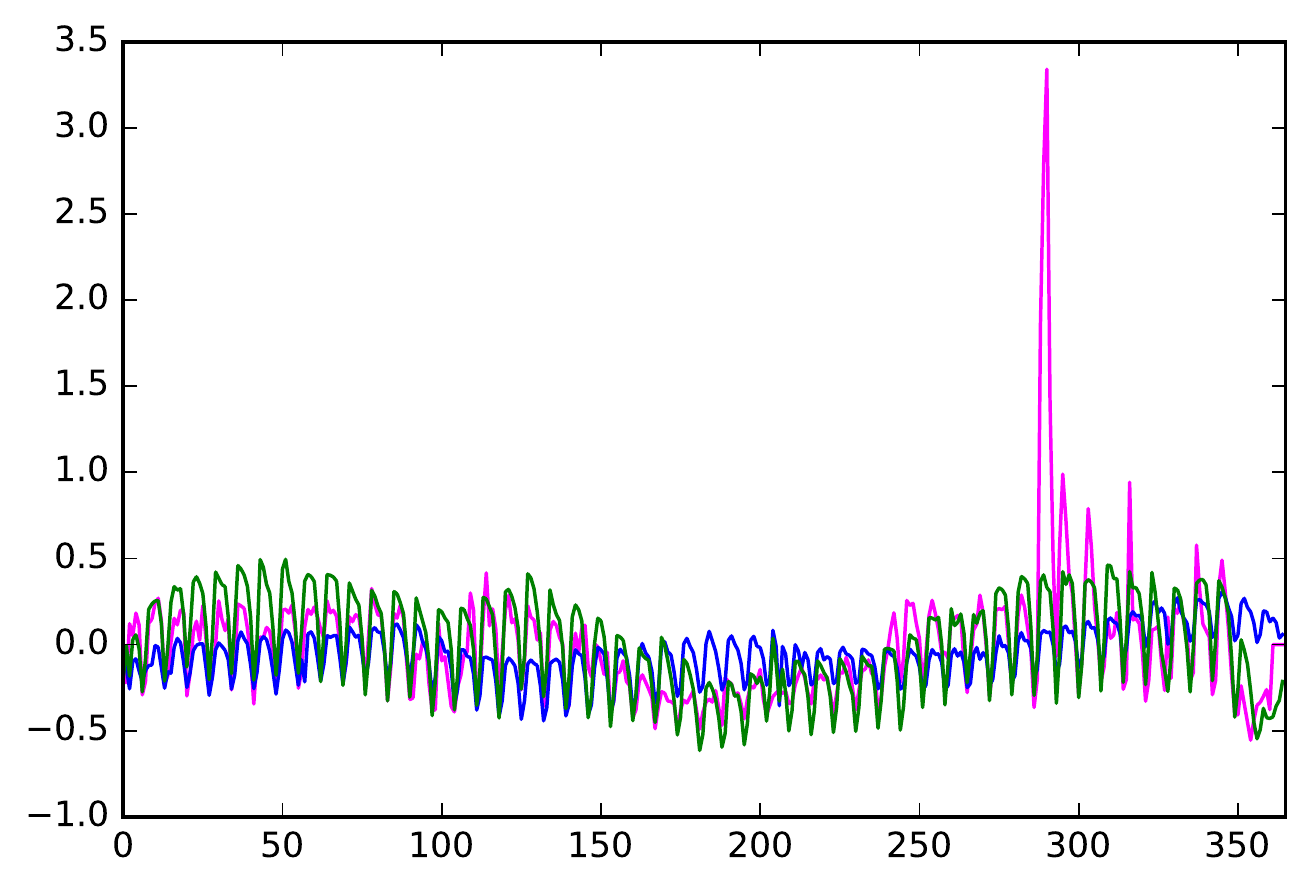}
\vspace{-.5cm}
\caption{Tutankhamun 2014}
\label{fig:trmf_comp_b}
\end{subfigure}
\hfill
\begin{subfigure}[t]{0.31\textwidth}
\includegraphics[width=\linewidth, height=2.5cm]{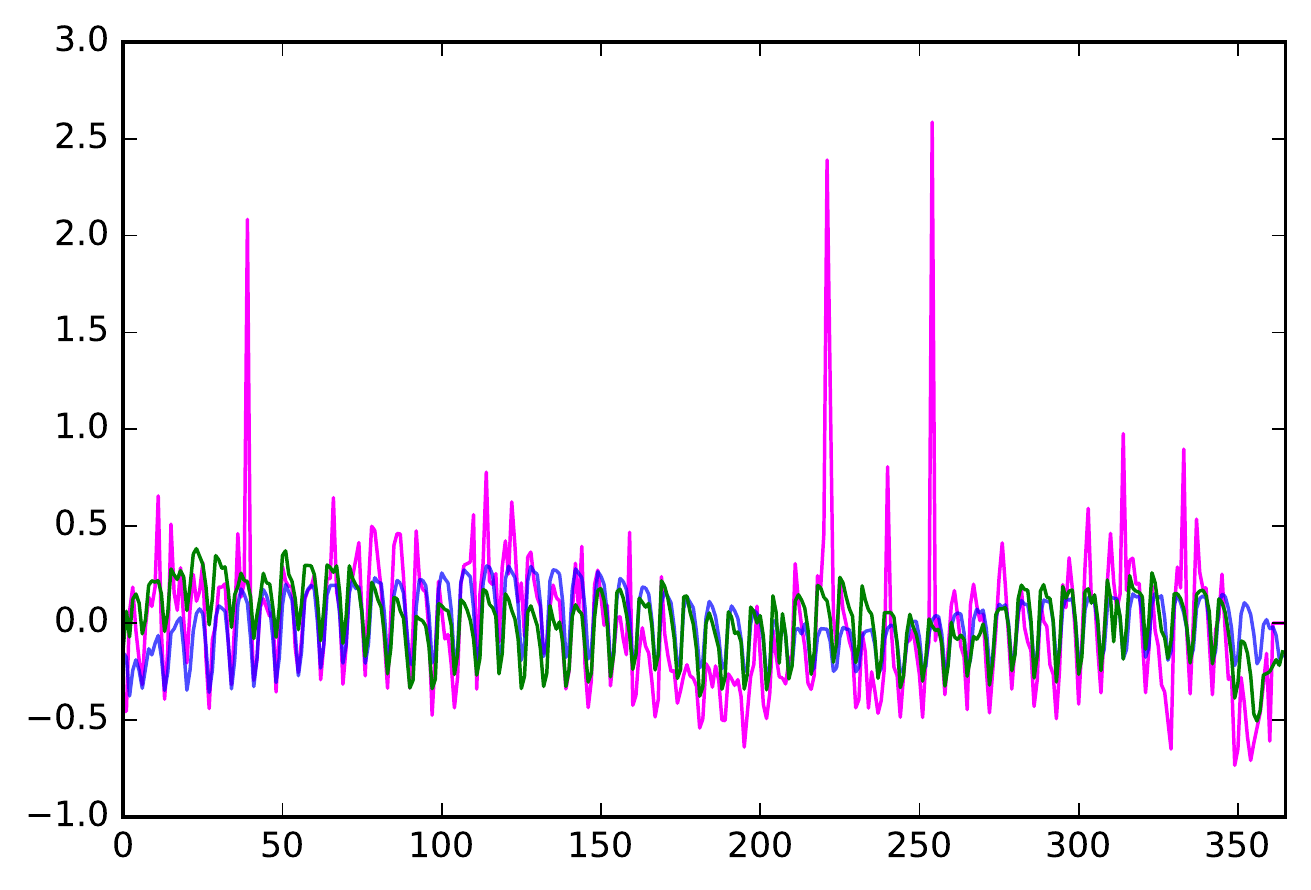}
\vspace{-.5cm}
\caption{British Raj 2014} 
\label{fig:trmf_comp_c}
\end{subfigure}

\vspace{-.2cm}
\caption{Comparison of the low-rank plus \texttt{MF} model and \texttt{TRMF} on the \texttt{Wiki} dataset for the long-range prediction challenge. The \textcolor{magenta}{observed time series} is in magenta, \textcolor{ForestGreen}{low-rank + \texttt{MF} predictions} are in green, and \textcolor{blue}{\texttt{TRMF}} predictions are in blue. Best viewed in color on a computer screen.}
\label{fig:trmf_comparison}
\end{center}
\end{figure*}

Since low-rank plus \texttt{MF} appears to be a fairly robust general purpose model, we hone in on this model when visually examining the performance of our predictions. We show that low-rank plus \texttt{MF} model is able to accurately forecast very intricate seasonal profiles over the course of an entire year both when provided with a past history of the series, and for completely unseen series. To show this, we plot selected time series from the \texttt{Wiki} test set and our associated predictions, and provide comparisons to the state-of-the-art \texttt{TRMF} baseline. See the appendix for more plots.

\prepar
\paragraph{Long-range forecasts} For the long-range forecasting challenge, our model captures a variety of interesting yearly patterns. Figures \ref{fig:longrange_a} displays a weekly oscillation and a summer and winter slump. We attribute this to students visiting this article during the school year (this specific yearly structure is common to many time series in the dataset, which is exploited by our low-rank regression and MF components). Figure \ref{fig:longrange_b} shows our model capturing an interesting seasonal pattern of an article corresponding to a TV series. Two modes of seasonality are captured: the weekly oscillation associated with weekly episodes, and the summer dip which is due to a summer break in the TV show. Looking at Figure \ref{fig:longrange_c} for ``Cherry Blossom'', we see predicted increases in page traffic during the months of the year when cherry blossoms are in full bloom.

\prepar
\paragraph{Cold-start forecasts} In the more challenging cold-start scenario, the model still excels at learning the common weekly oscillation with summer and winter slumps, when appropriate, as shown in Figure \ref{fig:pure_a} (the seasonal structure is similar to Figure \ref{fig:longrange_a}). Interestingly, we capture a slightly different type of weekly oscillation in Figure \ref{fig:pure_b}: this structure differs from Figures \ref{fig:longrange_a} and \ref{fig:pure_a} in that the beginning of the time series exhibits an upwards trend as opposed to an arch. This may be due to a different set of students (perhaps in a different part of the world) visiting these webpages, thus reflecting a different school schedule. In Figure \ref{fig:pure_c}, the model predicts increases in page traffic around days 75 to 100, which is most likely due to March Madness (annual college basketball tournament), in which the Villanova Wildcats participate in yearly. Although our model has never seen any of these series before, it is still capable of accurately forecasting these complex seasonal structures.

\paragraph{Comparison to \texttt{TRMF}} In Figure \ref{fig:trmf_comparison}, we plot comparisons of low-rank plus \texttt{MF} against the state-of-the-art \texttt{TRMF} on the long-range challenge. We selected series from the top 10\% of \texttt{TRMF} predictions according to our APST\_MSE metric (Eq. \ref{eq:per_series_thresholded_MSE}). In Figure \ref{fig:trmf_comp_a}, it is clear that our model more appropriately captures the bumps around days 100 and 160, and \texttt{TRMF} falsely predicts a weekly oscillation structure. Looking closely at Figure \ref{fig:trmf_comp_b}, both models capture the weekly oscillating structure, however our model more aptly predicts the scale and yearly pattern (i.e. assumed school schedule structure). Both models perform similarly on Figure \ref{fig:trmf_comp_c}. Despite constraining our visualization to where \texttt{TRMF} performs well, \texttt{TRMF} seems to fail at capturing essential seasonal structure, whereas our model does not. In general, \texttt{TRMF} seems to capture a more limited range of seasonal variation in comparison to our model. 


\postsec

\section{Conclusion and Future Work}
\presec
We have presented a computationally efficient forecasting framework applicable to a wide variety of scenarios: long-range, cold-start and warm-start settings. Our framework consists of a regression component that leverages high-dimensional metadata, and a matrix factorization term that exploits low-rank structure arising from shared patterns across fixed periods and series. Key to the framework is a clever re-organization of the data matrix that contributes to the low-dimensional shared seasonality structure.  

We explored different approaches to the regression component and examined the proposed formulations on Google Flu Trends and a large, messy Wikipedia dataset. Our experiments demonstrated that our framework excels at capturing intricate seasonal structure and can produce accurate long-range forecasts, even for unobserved series. Such forecasts are extremely challenging for standard time series models that perform iterated step-ahead predictions, leading to large error accumulation, especially under model misspecification. Our framework, on the other hand, side steps this issue by focusing prediction on the entire seasonal profile.  

There are many interesting directions for future work. First, we plan to investigate how our framework can be augmented to predict other time series features, such as trends and spikes. Both could be partially addressed by using time-varying metadata.  Second, one could leverage ideas from trend filtering to allow for learned latent factors to exhibit varying levels of smoothness, which would be more descriptive than our current functional regression approach.  Finally, it is interesting to explore the performance of our framework on predicting other types of multivariate functional output data from text, such as images based on captions.

\appendix

\section{Experiment Details}
\label{appendix:experiment_details}

\paragraph{Cross validation}
For each forecasting experiment, we separately run 5-fold cross validation over a grid in order to select the model hyperparameters. Because some of our loss functions are nonconvex, this requires running random restarts. Since this is computationally expensive, we perform an approximation to full grid search. We perform a two-stage cross validation where we first run standard 5-fold cross validation over the hyperparameters in the non-matrix factorization models. The regularization parameters for these models are a subset of the respective corresponding model that includes matrix factorization, so we then fix those hyperparameter values in the corresponding models with matrix factorization and cross validate over the remaining hyperparameters. Looking at Eq. (4) in the main paper, we would first cross validate over $\lambda_1$ and fix the selected value while cross validating over $\lambda_2$.

\subsection{Dataset Collection Details}

\subsubsection{Google Flu Trends}
The Flu Trends dataset consists of weekly estimates of influenza rates in 311 worldwide regions \cite{ginsberg2009detecting}. We scraped all of the data on the existing webpage, resulting in a matrix of 311 time series of weekly influenza rate estimates ($T=52$). For each time series, we zero-mean the data and divide by the standard deviation to standardize the scale. After reorganizing the matrix as described in Section 3, there are 3197 observed years of data. For metadata, we scraped the summary of the relevant Wikipedia page for each region. We removed stopwords, performed tokenization using the Stanford CoreNLP toolkit \cite{manning-EtAl:2014:P14-5}, and calculated TF-IDF representations of these summaries to use as our metadata vectors. Additionally, each word used in our feature vector must be present in at least two articles.  
After preprocessing, the dimension of the feature vector is $m = 3356$. This matrix is sparse with roughly $2.9\%$ nonzero entries.

\subsubsection{Wikipedia Page Traffic Dataset}
For the Wikipedia dataset, we collected daily page traffic counts from 4031 Wikipedia articles from the beginning of 2008 to the end of 2014 by querying Wiki Trends \cite{wikitrends}. The pages were selected by obtaining a list of the 5000 most popular pages of a given week in March 2016. 
We employ the method of \cite{cleveland1990stl} to detrend the time series data. For each time series, we zero-mean the data and divide by the standard deviation. For each article, we have anywhere between 1 to 7 years of page traffic counts starting from 2008 to 2014. After reorganizing the time series data matrix as described in Section 3 in the main paper, the total number of columns of $Y$ is $N = 29093$. Each year of article traffic has $T = 365$ days.   Furthermore, we shifted each year to start on the first Sunday and padded with zeros. We scraped the summary of the corresponding Wikipedia page, removed stopwords, performed tokenization using the Stanford CoreNLP toolkit \cite{manning-EtAl:2014:P14-5}, and calculated TF-IDF representations of these summaries to use as our metadata vectors. Additionally, each word used in our feature vector must be present in at least two articles. After preprocessing, the dimension of the features is $m = 22193$. This matrix is extremely sparse with roughly $0.5\%$ nonzero entries.

\subsection{Real Data Experiment Details}

\removed{
\paragraph{Test set selection} 
\begin{itemize}
\item[P1)] \textbf{Long-Range Forecasting}: We included the last year of page traffic for each article in the test set. We further removed 20\% of values from the training data to demonstrate the efficacy of our prediction methods in the presence of missing data.

\item[P2)] \textbf{Missing Data}: For each time series, we pick a time point uniformly from the year for the starting point of the missing chunk, and sample a geometric random variable with mean $\frac{T}{2}$ that determines the length of the missing chunk.

\item[P3)] \textbf{Cold-Start Forecasting}: We randomly selected 25\% of the articles and included the last year of page traffic in the test set. We did not include any previous years of such articles from the training set. We further removed 20\% of values from the training data to demonstrate the efficacy of our prediction methods in the presence of missing data.

\end{itemize}
}

\paragraph{Neural network architecture} For the neural network regression, we use a two hidden layer network with ReLU activations. The first hidden layer has 100 nodes and the second hidden layer has $T = 365$ nodes. Finally, the output layer is a regression layer (uses the identity function as the activation function). For simplicity we also do not regularize the weights of the neural network.

\paragraph{Model training: Google Flu} We set $k = k' = 5$ and train the models with minibatch stochastic gradient descent (SGD). We use a minibatch size of 300 and 30,000 iterations. We individually tuned constant step sizes for each model, although they all turned out to be close to 1e-2.  We use 5-fold cross validation to select hyperparameters. For both $\lambda_1$ and $\lambda_2$, we select them from 10 values evenly spread on a log scale from 0.1 to 1000. For the functional regression models, we choose the number of (evenly spaced) knots $K$ from 4, 8, ..., 20. After selecting hyperparameters, we trained each final model with 50,000 iterations and 10 random restarts. We use a validation set to tune the number of epochs the neural network regression models need.

\paragraph{Model training: Wikipedia Page Traffic} We set $k = k' = 20$ and train the models with minibatch SGD. We use a minibatch size of 500 and 10,000 iterations. We individually tuned constant step sizes for each model, although they all turned out to be close to 1e-1 (with the exception that the neural network regression models used a step size of 1e-2). In this setting, since our data is large enough, we use a single validation set to select our hyperparameters as opposed to cross validation. For both $\lambda_1$ and $\lambda_2$, we select them from 10 values evenly spread on a log scale from 0.01 to 1000. We choose the number of (evenly spaced) knots $K$ from $10, 20, \ldots, 100$. After selecting hyperparameters, we trained each final model with 20,000 iterations and 10 random restarts. We use a validation set to tune the number of epochs the neural network regression models need.

\paragraph{Baselines} For TRMF \cite{Yu2015highdim}, we fix latent dimension ($k$ in our framework) to be the same as what we set for our framework (5 for Google Flu, 20 for the Wikipedia dataset). For both TRMF and the AR baseline, we use a lag set of $\{ 1, \ldots, 5 \} \cup \{ 51, 52, 53 \}$ for the Google Flu data, and a lag set of $\{ 1, \ldots, 10 \} \cup \{ 63, \ldots, 70 \} \cup \{ 363, 364, 365, 366\}$. The chosen lag sets allow for the step-ahead models to learn seasonal yearly patterns. For TRMF, we run rolling cross validation to select hyperparameters, using the code provided by \cite{Yu2015highdim}, while for AR, we simply perform maximum likelihood under Gaussian errors. For \texttt{MF} alone, we run cross validation to select the hyperparameters for regularization. For the \texttt{$k$-NN} baseline, we compared $k = 5, 10, 15, 20$ and selected $k = 10$ due to best performance.


\newpage
\bibliography{cold_start_bib}

\begin{thebibliography}{10}

\bibitem{wen:2017}
Ruofeng Wen, Kari Torkkola, and Balakrishnan Narayanaswamy.
\newblock A multi-horizon quantile recurrent forecaster.
\newblock {\em arXiv preprint arXiv:1711.11053}, 2017.

\bibitem{yu:2017}
Rose Yu, Stephan Zheng, Anima Anandkumar, and Yisong Yue.
\newblock Long-term forecasting using tensor-train {RNN}s.
\newblock {\em arXiv preprint arXiv:1711.00073}, 2017.

\bibitem{lutkepohl:2005}
Helmut L{\"u}tkepohl.
\newblock {\em New introduction to multiple time series analysis}.
\newblock Springer Science \& Business Media, 2005.

\bibitem{xia2011}
Yingcun Xia and Howell Tong.
\newblock Feature matching in time series modeling.
\newblock {\em Statist. Sci.}, 26(1):21--46, 02 2011.

\bibitem{seeger2016bayesian}
Matthias~W Seeger, David Salinas, and Valentin Flunkert.
\newblock Bayesian intermittent demand forecasting for large inventories.
\newblock In D.~D. Lee, M.~Sugiyama, U.~V. Luxburg, I.~Guyon, and R.~Garnett,
  editors, {\em Advances in Neural Information Processing Systems 29}, pages
  4646--4654. Curran Associates, Inc., 2016.

\bibitem{nguyen2014collaborative}
Trung~V Nguyen and Edwin~V Bonilla.
\newblock Collaborative multi-output gaussian processes.
\newblock In {\em UAI}, pages 643--652, 2014.

\bibitem{sun2015time}
Wei Sun and Dmitry Malioutov.
\newblock Time series forecasting with shared seasonality patterns using
  non-negative matrix factorization.
\newblock In {\em NIPS, Time Series Workshop}, 2015.

\bibitem{Li:2015}
Steve Cheng-Xian Li and Benjamin Marlin.
\newblock Collaborative multi-output gaussian processes for collections of
  sparse multivariate time series.
\newblock {\em NIPS 2015 Time Series Workshop}, 2015.

\bibitem{Yu2015highdim}
Hsiang-Fu Yu, Nikhil Rao, and Inderjit~S Dhillon.
\newblock Temporal regularized matrix factorization for high-dimensional time
  series prediction.
\newblock In {\em Advances in Neural Information Processing Systems}, pages
  847--855, 2016.

\bibitem{de2008analysis}
Ruair{\'\i} de~Fr{\'e}in, Konstantinos Drakakis, Scott Rickard, and Andrzej
  Cichocki.
\newblock Analysis of financial data using non-negative matrix factorization.
\newblock In {\em International Mathematical Forum}, volume~3, pages
  1853--1870. Journals of Hikari Ltd, 2008.

\bibitem{agarwal:2017}
Anish Agarwal, Muhammad~Jehangir Amjad, Devavrat Shah, and Dennis Shen.
\newblock Time series forecasting via matrix estimation.

\bibitem{ramsay2005functional}
James~O.. Ramsay and Bernard~W Silverman.
\newblock {\em Functional Data Analysis}.
\newblock Springer, 2005.

\bibitem{morris2014functional}
J.~S. {Morris}.
\newblock {Functional Regression}.
\newblock {\em ArXiv e-prints}, June 2014.

\bibitem{koren2009matrix}
Yehuda Koren, Robert Bell, and Chris Volinsky.
\newblock Matrix factorization techniques for recommender systems.
\newblock {\em Computer}, 42(8), 2009.

\bibitem{agarwal2009regression}
Deepak Agarwal and Bee-Chung Chen.
\newblock Regression-based latent factor models.
\newblock In {\em Proceedings of the 15th ACM SIGKDD international conference
  on Knowledge discovery and data mining}, pages 19--28. ACM, 2009.

\bibitem{pilaszy2009recommending}
Istv{\'a}n Pil{\'a}szy and Domonkos Tikk.
\newblock Recommending new movies: even a few ratings are more valuable than
  metadata.
\newblock In {\em Proceedings of the third ACM conference on Recommender
  systems}, pages 93--100. ACM, 2009.

\bibitem{schein2002methods}
Andrew~I Schein, Alexandrin Popescul, Lyle~H Ungar, and David~M Pennock.
\newblock Methods and metrics for cold-start recommendations.
\newblock In {\em Proceedings of the 25th annual international ACM SIGIR
  conference on Research and development in information retrieval}, pages
  253--260. ACM, 2002.

\bibitem{gantner2010learning}
Zeno Gantner, Lucas Drumond, Christoph Freudenthaler, Steffen Rendle, and Lars
  Schmidt-Thieme.
\newblock Learning attribute-to-feature mappings for cold-start
  recommendations.
\newblock In {\em Data Mining (ICDM), 2010 IEEE 10th International Conference
  on}, pages 176--185. IEEE, 2010.

\bibitem{goodfellow2016deep}
Ian Goodfellow, Yoshua Bengio, and Aaron Courville.
\newblock {\em Deep Learning}.
\newblock MIT Press, 2016.
\newblock \url{http://www.deeplearningbook.org}.

\bibitem{ginsberg2009detecting}
Jeremy Ginsberg, Matthew~H Mohebbi, Rajan~S Patel, Lynnette Brammer, Mark~S
  Smolinski, and Larry Brilliant.
\newblock Detecting influenza epidemics using search engine query data.
\newblock {\em Nature}, 457(7232):1012--1014, 2009.

\bibitem{cleveland1990stl}
Robert~B Cleveland, William~S Cleveland, Jean~E McRae, and Irma Terpenning.
\newblock Stl: A seasonal-trend decomposition procedure based on loess.
\newblock {\em Journal of Official Statistics}, 6(1):3--73, 1990.

\bibitem{manning-EtAl:2014:P14-5}
Christopher~D. Manning, Mihai Surdeanu, John Bauer, Jenny Finkel, Steven~J.
  Bethard, and David McClosky.
\newblock The {Stanford} {CoreNLP} natural language processing toolkit.
\newblock In {\em Association for Computational Linguistics (ACL) System
  Demonstrations}, pages 55--60, 2014.

\bibitem{wikitrends}
Wikipedia.
\newblock Wikipedia pageview statistics: Wiki trends.
\newblock \url{http://www.wikipediatrends.com/}, 2016.

\end{thebibliography}
\bibliographystyle{unsrt}

\end{document}